\documentclass[letterpaper,journal]{IEEEtran}
\usepackage{amsmath,amsfonts,amssymb}
\usepackage{algorithmic}
\usepackage{algorithm}
\usepackage{array}
\usepackage[caption=false,font=normalsize,labelfont=sf,textfont=sf]{subfig}
\usepackage{textcomp}
\usepackage{stfloats}
\usepackage{url}
\usepackage{verbatim}
\usepackage{graphicx}
\usepackage{cite}
\begin{document}

\title{Bridging Semantics and Physical Execution: A Neuro-Symbolic Framework for Multi-Pair Robotic Assembly}

\author{Xinyi Li, Aiguo Song\textsuperscript{*}, Linhu Wei, and Huijun Li
\thanks{School of Instrument Science and Engineering, Southeast University, Nanjing 210096, China.}%
\thanks{*Corresponding author: Aiguo Song. Email: a.g.song@seu.edu.cn}}

\maketitle

\begin{abstract}
Multi-pair robotic assembly in unstructured environments faces spatial interference and contact uncertainties. We propose a neuro-symbolic framework that addresses the challenge hierarchically: generating local task graphs for each pair, decoupling generality from particularity, and then resolving cross-pair interferences. Given an eye-on-hand RGB-D assembly scene, the framework extracts semantic instance identities and state probabilities. For each pair, a local graph is generated via a large language model (LLM) using only basic assembly actions. Supportive actions for edge cases are inferred and inserted by a lightweight discriminator. Driven by the divergence between a canonical reference and the current task, it supports knowledge reuse across assembly structures and supervised extension to new actions. Augmented subgraphs are topologically coordinated into global sequences while preserving internal behavioral coherence. Dynamic behavior trees embedding atomic skills enable force-aware execution with feedback within individual actions. Offline evaluation on 100 scenes achieves 95\% strict complete-scene success, matching the tested hierarchical task network (HTN) baseline and outperforming the other evaluated planners. Real-robot deployment on a UR3 arm achieves 57/60 successes with a maximum radial clearance of 0.5~mm, demonstrating a practical solution for complex autonomous assembly.
\end{abstract}


\section{Introduction}

\IEEEPARstart{R}{obotic} assembly has long been a cornerstone of intelligent manufacturing and flexible automation. However, scaling from rigid single-object operations with fixed scripts to long-horizon multi-pair assembly in unstructured environments remains an open challenge~\cite{ranjan_das_toward_2025,karbouj_adaptive_2026,wang_llm-based_2025,zhang_surrogate_2025}. Such tasks involve varying initial states, spatial constraints, and contact uncertainties inherent in tight-fitting tasks such as gear-on-shaft and peg-in-hole assembly, as highlighted by standardized benchmarking studies~\cite{van_wyk_comparative_2018}. Reliable and efficient execution demands three core capabilities: (i) a robust perception layer that explicitly quantifies semantic uncertainty; (ii) a high-level planner that suppresses logical hallucinations while preserving commonsense flexibility and global assembly efficiency; and (iii) a force-aware execution layer that grounds symbolic plans in physical reality. Moreover, for complex geometric dependencies such as object stacking, globally valid assembly sequence planning is essential not only for efficiency but also for physical reachability.

Bridging cognitive decision-making and physical execution remains challenging in multi-pair assembly. Classical symbolic planners support verification of topological correctness within the modeled domain. PDDL~\cite{fox_pddl21_2003} specifies planning domains and problems, while HTN planners~\cite{nau_shop2_2003} organize procedures through predefined task-decomposition methods, requiring effort to encode domain knowledge. Recent task and motion planning (TAMP) architectures mitigate grounding gaps via execution-time feedback~\cite{pan_task_2024}, although their capacity for commonsense adaptation in unstructured environments may remain limited. In contrast, large language models (LLMs) offer commonsense task decomposition from semantic instructions, but generated action sequences can contain logical hallucinations and topological conflicts in long-horizon planning~\cite{liu_delta_2025}. Even logically feasible plans stay fragile during contact-rich insertion with tight tolerances, motivating the integration of discrete symbolic action sequences with continuous force-aware control. These complementary strengths and limitations motivate combining learned components with explicit symbolic models, as in neuro-symbolic planning~\cite{sun_neurosymbolic_2024}, to integrate flexible task reasoning, constraint checking, and force-aware execution for long-horizon multi-pair robotic assembly.

\begin{figure}[!t]
\centering
\includegraphics[width=\columnwidth]{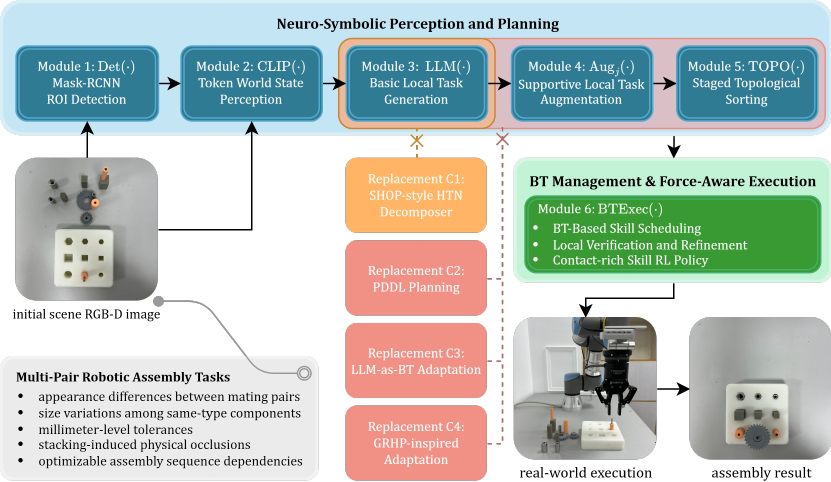}
\caption{The neuro-symbolic assembly pipeline and replacement scopes of the comparison methods C1--C4. This figure demonstrates the workflow from pixel-level observation $I_t$ to force-aware execution. Dashed lines indicate the planning components replaced by each method.}\label{fig:pipeline_overview}
\end{figure}

To fill this gap, we propose a neuro-symbolic framework with a local-global hierarchical design, illustrated in Fig.~\ref{fig:pipeline_overview}, bridging semantic probability distributions, topological planning, and force-aware execution. The modular planning pipeline generates local task graphs, augments them with supportive actions, and coordinates them into global execution sequences. Physical execution uses dynamic behavior trees (BTs), force-feedback reinforcement learning (RL) policies for contact-rich processes, and other low-level skills. In assembly scenes covering multiple assembly types, different initial states, and diverse stacking dependencies, our framework matches the tested HTN baseline~\cite{nau_shop2_2003} and outperforms the other evaluated planners~\cite{fox_pddl21_2003,ao_llm-as-bt-planner_2025,li_grhp_2026} in complete-scene success.

Specifically, our framework decouples long-horizon assembly logic into an extensible hierarchy of basic actions and supportive actions. Common assembly logic is abstracted as basic actions, which the LLM generates directly from structured states. Sparse yet critical cases—such as occlusion clearance, occupancy release, and pose alignment—are modeled as pluggable supportive actions. A lightweight Jensen-Shannon divergence-driven discriminator (JS-Aug) uses semantic distribution differences between a canonical reference and variant scenes to infer supportive-action requirements, guiding augmentation of the basic-action graphs. This decoupling preserves basic assembly logic generality and LLM flexibility while handling scene-dependent supportive requirements explicitly. Consequently, the resulting discriminator supports knowledge reuse across assembly structures and supervised extension to new actions.

The core contributions of this paper are threefold:
\begin{enumerate}
\item We devise JS-Aug, a lightweight discriminator that decouples supportive-action reasoning for edge cases from LLM basic-action generation. Trained with limited supervision, it infers supportive-action requirements from semantic differences relative to a canonical reference, supporting knowledge reuse across assembly structures and supervised extension to new actions.
\item We propose a neuro-symbolic framework that converts visual assembly scenes into global action sequences \(\sigma_t\) through a local-global graph hierarchy. LLM-generated basic-action subgraphs are augmented with supportive actions and coordinated through cross-pair dependency inference and action ordering to resolve dependencies such as stacking, prune redundant actions, and preserve local assembly logic.
\item We bridge neuro-symbolic planning and physical execution through dynamic behavior-tree scheduling of atomic skills with local validation and correction. Modular cache replay evaluates perception and planning components, while real-robot trials demonstrate the feasibility of integrating coordinated plans with force-aware execution in the tested assembly tasks.
\end{enumerate}

\section{Related Work}
Robotic assembly is shifting from rigid scripted pipelines toward hybrid intelligent systems. Reliable autonomous execution in unstructured environments faces challenges in semantic understanding, task coordination, and physical control.

\subsection{Multi-pair Robotic Assembly}
Multi-pair assembly tasks, such as aircraft panel docking and multi-component disassembly of retired products, require coordinated execution of peg-in-hole, gear-shaft, and nut-bolt operations under stacking dependencies, shared resources, and dynamic disturbances. Recent reviews~\cite{ranjan_das_toward_2025,karbouj_adaptive_2026} highlight the difficulty of scaling single-object or rigidly structured solutions under environmental and human variability.

Classical assembly-sequence planning represents assembly alternatives with AND/OR graphs, component connections with liaison relations, and assembly order with precedence constraints~\cite{homem_de_mello_andor_1990,jimenez_survey_2013}. Given assembly pairs, our framework uses scene semantics to construct, augment, and coordinate local action graphs. In high-level planning, LLM-based multi-agent frameworks~\cite{wang_llm-based_2025,keramat_decentralized_2026} use genetic algorithms or decentralized oracles to decompose natural-language intents while balancing operator experience and efficiency. LLM-generated sequences still require validation of cross-pair dependencies to guard against logical hallucinations. Symbolic planners such as HAD-TAMP~\cite{gottardi_had-tamp_2026} enforce modeled logical constraints and support dynamic replanning, while planning cost and perceptual uncertainty remain important considerations.

For multi-component assembly coordination, surrogate-model-driven MARL frameworks~\cite{zhang_surrogate_2025} effectively co-optimize component deformation and inter-panel clearances in large-scale parallel assembly. Nevertheless, extending surrogate-based geometric coordination to contact-rich multi-pair assembly also requires handling scene-dependent action requirements and execution dependencies. At the force-aware execution layer, meta-RL approaches~\cite{chen_behavioral_2026} achieve rapid parameter adaptation via PEARL-SAC under environmental interference. Together, these approaches motivate linking task planning and cross-pair dependency handling with force-aware execution.

\subsection{Instance Perception and VLM in Robotic Manipulation}

Reliable instance-level perception is a prerequisite for multi-pair assembly, since low-level semantic uncertainties (e.g., physical occlusions, rotational ambiguity, and grasping states) readily propagate errors into high-level planning. Recent works~\cite{kawaharazuka_vision-language-action_2025,liu_vla-pruner_2026,wang_vlabot_2026} in vision-language models (VLMs/VLAs) demonstrate semantic understanding, efficient inference through temporal-aware visual token pruning, and natural human-robot interaction via augmented reality for long-horizon assembly.

VLM-DEWM~\cite{tang_vlm-dewm_2026} decouples VLM reasoning from persistent state tracking to handle out-of-view objects and state drift, while~\cite{chen_generalizable_2026} combines LLM-guided ontology with geometric similarity matching for generalizable task-oriented grasping on novel objects. Complementary efforts address error mitigation through closed-loop correction and human-in-the-loop feedback~\cite{zhou_genco_2025,zhou_human---loop_2025}.

These studies advance semantic grounding and state tracking. For downstream planning, semantic ambiguity motivates explicit representation of scene semantics and uncertainty.

\subsection{The Spectrum of Neuro-Symbolic Planning}
Traditional TAMP combines symbolic solvers and motion planning for constraint validation, requiring appropriate task models and perceptual grounding in unstructured assembly~\cite{garrett_integrated_2021}. Partial-order planners such as SNLP and UCPOP use causal links and ordering constraints to protect action preconditions~\cite{mcallester_systematic_1991,penberthy_ucpop_1992}. TOPO applies heuristic dependency handling and action ordering to augmented local assembly graphs.

Recent neuro-symbolic approaches combine semantic reasoning with BTs and hierarchical structures for flexibility and reactivity. Works such as~\cite{zhou_llm-bt_2024,ao_llm-as-bt-planner_2025,styrud_automatic_2025,izzo_btgenbot_2024} leverage LLMs for direct BT generation or dynamic expansion, enabling adaptive execution. HBTP~\cite{cai_hbtp_2025} introduces LLM-guided heuristic BT planning, Kwon et al.~\cite{kwon_fast_2025} employ multi-level goal decomposition, and GRHP~\cite{li_grhp_2026} uses graph-fused architectures to bridge high-level semantics with low-level execution. Merlo et al.~\cite{merlo_exploiting_2025} use information-theoretic interaction representations and scene graphs to derive behavior trees from demonstrations. SayCan~\cite{ichter_do_2023} grounds language-based skill selection in learned affordances. ProgPrompt~\cite{singh_progprompt_2023} and Code as Policies~\cite{liang_code_2023} generate programs using robot actions and perceptual information.

LLM+P~\cite{liu_llmp_2023} delegates planning to symbolic solvers, while LLM-Modulo~\cite{kambhampati_llm_modulo_2024} couples LLM proposals with external verification. Mendez-Mendez~\cite{mendez-mendez_systematic_2025} reports lower success rates for the tested LLM-based TAMP variants than for their engineered counterparts, while works such as~\cite{shao_breaking_2025,bhat_grounding_nodate,rodriguez-guerra_deliberative_2025,galitsky_neuro-symbolic_2026} introduce symbolic verifiers, closed-loop feedback, deliberative layered BTs, and abductive reasoning to mitigate hallucinations and constraint violations. Within this spectrum, our framework offers a complementary perspective by separating common task logic from scene-specific requirements during plan construction.

\subsection{Force-Aware Control in Contact-Rich Manipulation}
Contact-rich manipulation, particularly high-precision peg-in-hole and connector insertion, is challenging due to frictional dynamics, pose uncertainty, and tight clearances. Recent surveys~\cite{parnada_towards_2026,xie_towards_2025} highlight the promise and persistent cost and safety barriers of learning-based methods for such tasks.

Liao et al.~\cite{liao_friction-aware_2026} develop friction-aware strategies via real-time contact wrench measurement, while Tracy et al.~\cite{tracy_efficient_2025} propose efficient online learning of quasi-static contact force models for connector insertion. Works such as~\cite{hou_force-based_2025,shen_learning-based_2025} further demonstrate effective off-robot force sensing and time-series force-based state recognition for inclined-hole and large-scale assembly.

Ding et al.~\cite{ding_ensemble_2026} combine imitation learning with ensemble RL to achieve robust sim-to-real transfer in high-precision peg-in-hole tasks. Shirai et al.~\cite{shirai_sim--real_2025} integrate contact-implicit trajectory optimization with RL for pivoting manipulation, while SHaRe-RL~\cite{stranghoner_share-rl_2026} employs structured primitives with human-in-the-loop corrections for industrial assembly.

However, in multi-pair assembly, perception errors or logical hallucinations can produce physically infeasible plans, which local force control may not resolve without task-level handling of cross-pair dependencies.

\section{Problem Formulation}
This section formalizes the multi-pair assembly problem and presents the algebraic information flow of the proposed neuro-symbolic framework. A unified mathematical notation is used throughout this paper.

\subsection{Notation and Problem Statement}
At time $t$, the robot acquires an RGB-D observation image $I_t$ from an Eye-on-Hand camera. A detection and segmentation frontend extracts $N_t$ regions of interest (ROIs) from image $I_t$, forming an instance set:
\begin{equation}
\mathcal{R}_t = \{ r_{t,i} \}_{i=1}^{N_t}.
\end{equation}
Each instance $r_{t,i}$ maps to a scene object $i$ and contains ROI, category, and geometric information.

The unstructured nature of the assembly scene arises from (i) diversity in component types, quantities, and initial workspace states; (ii) complex spatial dependencies (e.g., stacking dependencies and physical occlusions) and contact uncertainties among components; and (iii) concurrent execution conflicts induced by shared resources (a single gripper and restricted workspace). Task objectives are issued in natural language and may specify attribute-paired assemblies (e.g., smallest circle peg-in-hole), same-category assemblies (e.g., square peg-in-hole), or macroscopic tasks (e.g., assembly). A multi-pair assembly task comprises $M_t$ local sub-task assembly pairs indexed by $j \in \{1, \dots, M_t\}$. Each pair $j$ is formalized as a triplet $\langle m_j, f_j, \mathrm{tools}_j \rangle$, where the male operative component $m_j$ (e.g., peg, gear) and female recipient component $f_j$ (e.g., hole, shaft) are dynamically matched from the instance set $\mathcal{R}_t$ according to geometric and semantic compatibility constraints.

For each pair $j$, the LLM first generates a local task DAG:
\begin{equation}
\pi_j = (V_j, E_j),
\end{equation}
where each node $v \in V_j$ is an action step of the form $(\mathrm{id},\mathrm{action},\mathrm{target},\mathrm{params})$ and directed edges $E_j$ represent precedence dependencies within the pair. The candidate action set consists of basic actions $\mathcal{A}_{\mathrm{base}} = \{\mathtt{localize}, \mathtt{pick}, \mathtt{insert}\}$ and supportive actions $\mathcal{A}_{\mathrm{supp}}=\{\mathtt{release},\mathtt{remove},\mathtt{align}\}$, extendable on demand.

The core system objective is to compute a globally executable discrete sequence $\sigma_t$ that satisfies (i) all internal and cross-pair topological precedence constraints, (ii) proactive handling of scene-related requirements via supportive actions, and (iii) grounding into a dynamic behavior tree while maintaining control robustness against millimeter-level contact uncertainties in the real world.

\subsection{Formal Objective and Algebraic Flow}
\begin{figure*}[!t]
\centering
\includegraphics[width=\textwidth]{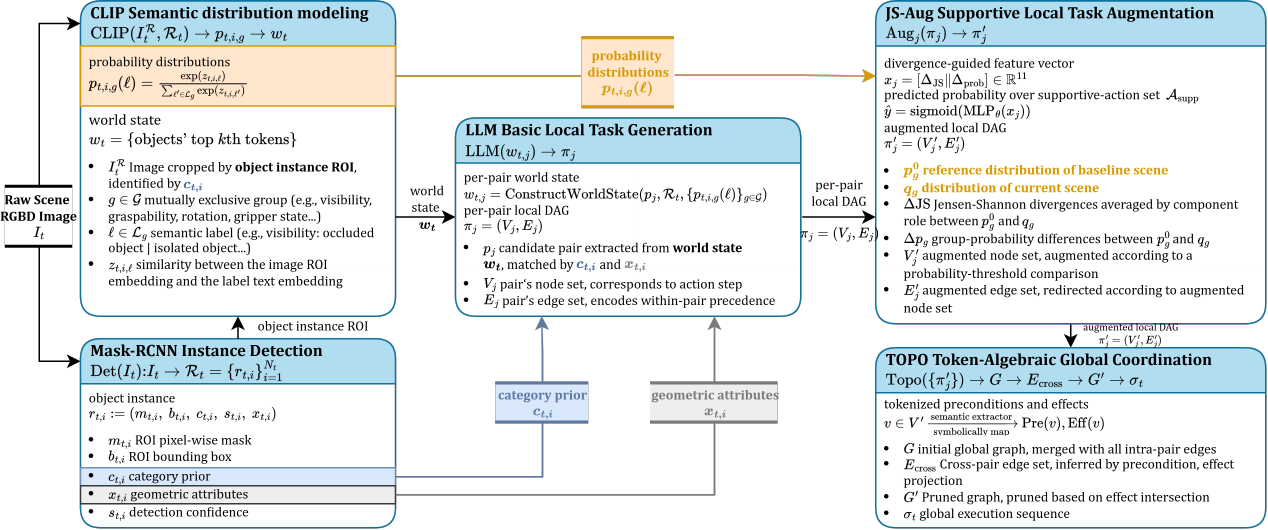}
\caption{Schematic of the system's algebraic information flow and token state transitions in multi-pair assembly scenes. The figure shows the complete pipeline from pixel observation and semantic probability distribution modeling through local DAG generation and augmentation to global token-based topological constraints.}
\label{fig:algebraic_flow}
\end{figure*}

To achieve the above objective, we construct a modular neuro-symbolic pipeline. The information flow is illustrated in Fig.~\ref{fig:algebraic_flow} and summarized by the following mappings:
\begin{equation}
\label{eq:infoflow}
\begin{split}
q_t &= \mathrm{BTExec}(\sigma_t),\\
\sigma_t &= \mathrm{Topo}\!\left(\left\{\mathrm{Aug}_j\!\left(\mathrm{LLM}\!\left(\mathrm{CLIP}\!\left(\mathrm{Det}(I_t)\right)\right)\right)\right\}_{j=1}^{M_t}\right).
\end{split}
\end{equation}

In the early stages of the information flow $\mathrm{Det} \rightarrow \mathrm{CLIP} \rightarrow \mathrm{LLM} \rightarrow \mathrm{Aug}$, the system performs local computations over individual ROIs or individual assembly pairs, progressively converting pixel-level uncertainty into ambiguity-resistant local symbolic DAGs. The TOPO stage is the first global coordination operator; it aggregates token-level precondition and effect constraints across all $M_t$ pairs and coordinates action ordering to resolve the conflicts. Finally, the BTExec module grounds the abstract global symbolic sequence $\sigma_t$ into continuous physical trajectories $q_t$ of the robotic arm, completing the perception-planning-execution pipeline with feedback within individual actions.

\setlength{\textfloatsep}{1.15\baselineskip plus 0.4\baselineskip}
\setlength{\floatsep}{0.65\baselineskip plus 0.4\baselineskip}
\setlength{\dbltextfloatsep}{1.15\baselineskip plus 0.4\baselineskip}
\section{Neuro-Symbolic Perception and Planning}

\subsection{Semantic State Graph Construction}
In long-horizon robotic assembly tasks, high-dimensional noisy raw images $I_t$ are transformed into structured symbolic states for high-level planners. The perception frontend uses a deep cascade of Mask R-CNN~\cite{he_mask_2017} and CLIP~\cite{radford_learning_2021} to model physical states from pixel observations and converts visual ambiguity and uncertainty into structured probability distributions serving as a unified interface for world-state construction $w_{t,j}$, local task graph generation, and global topological coordination.

\subsubsection{Instance Extraction and Context-Aware Encoding}
Mask R-CNN discretizes the input image to extract the instance set $\mathcal{R}_t=\{r_{t,i}\}_{i=1}^{N_t}$. Each instance contains a pixel-wise mask $m_{t,i}$, bounding box $b_{t,i}$, and initial detection category prior $c_{t,i}$.

For each instance $r_{t,i}$, adaptive ROI cropping expands the mask's bounding box into a square input for the vision-language encoder, yielding $\mathrm{crop}(I_t,m_{t,i})$. The vision encoder $f_\theta$ maps the processed ROI to a normalized image embedding:
\begin{equation}
u_{t,i}=\mathrm{normalize}\bigl(f_\theta(\mathrm{crop}(I_t,m_{t,i}))\bigr)
\end{equation}

On the text side, prompts are partitioned into \textbf{Identity} prompts directly injected from $c_{t,i}$ and \textbf{State} labels characterizing assembly-related attributes. Template ensembling generates multiple templated variants $\{\tau_{\ell}^{(m)}\}_{m=1}^{M_\ell}$ for each candidate label $\ell$. The text encoder extracts and normalizes each variant; the mean is taken and normalized again to obtain the aggregated text embedding:
\begin{equation}
\bar v_{\ell}=\mathrm{normalize}\Bigl(\tfrac{1}{M_\ell}\sum_{m=1}^{M_\ell} \mathrm{normalize}\bigl(g_\theta(\tau_\ell^{(m)})\bigr)\Bigr).
\end{equation}
The cosine similarity between instance $r_{t,i}$ and label $\ell$ in the shared latent space is scaled by $\exp(\alpha)$, where $\alpha$ is a learnable log-scale:
\begin{equation}
z_{t,i,\ell}=\exp(\alpha)\,\langle u_{t,i},\bar v_{\ell}\rangle.
\end{equation}

\subsubsection{Structured Prompt Engineering and Probabilistic Constraint Modeling}
Mutually exclusive semantic groups $\mathcal{G}$ address probability dispersion and logical conflicts in multimodal reasoning. Using the detection category prior $c_{t,i}$, domain-knowledge filtering removes labels inconsistent with the entity's physical properties (e.g., pose alignment is evaluated only for ``hole'' entities). For each group $g\in\mathcal{G}$ and its active candidate label subset $\mathcal{L}_g$, the intra-group conditional probability distribution $p_{t,i,g}(\ell)$ is computed:
\begin{equation}
\label{eq:clip_group_prob}
p_{t,i,g}(\ell) = \frac{\exp(z_{t,i,\ell})}{\sum_{\ell' \in \mathcal{L}_g} \exp(z_{t,i,\ell'})},
\end{equation}

The detector prior $c_{t,i}$ filters the prompt set to retain only category-consistent active labels. Normalization is performed exclusively within these active labels. This converts similarity scores into model-estimated physical-state probabilities.

Discretized group-value encodings are provided for critical mutually exclusive groups (e.g., visibility state, grasping state) to support downstream planning and execution checks. This yields the discrete semantic state $\ell^*_{t,i,g}=\arg\max_{\ell\in\mathcal{L}_g}p_{t,i,g}(\ell)$ for each group. Classification criteria are detailed in Table~\ref{tab:clip_groups}.

\begin{table*}[!t]
\caption{CLIP Prompt Classification and Critical Mutually Exclusive Semantic Groups}
\label{tab:clip_groups}
\centering
\renewcommand{\arraystretch}{1.15}
\begin{tabular}{llll>{\raggedright\arraybackslash}p{0.6\textwidth}}
\hline
\textbf{Prompt} & \textbf{Semantic} & \textbf{Filtered} & \textbf{Group} & \textbf{Example} \\
\textbf{Type} & \textbf{Group $g$} & \textbf{by $c_{t,i}$} & \textbf{Value} & \textbf{Prompts} \\
\hline
 & peg\_identity & peg & --- & \texttt{a \{peg\_type\} peg} \\
 & hole\_identity & hole & --- & \texttt{a \{hole\_type\} hole} \\
Identity & gear\_identity & gear & --- & \texttt{a \{gear\_type\}} \\
 & gripper\_identity & gripper & --- & \texttt{a gripper}; \texttt{two-finger gripper}; \texttt{parallel gripper} \\
\hline
 & visibility & peg, gear & 1/0 & 1: \texttt{single object}; \texttt{one object visible}; \texttt{isolated object}.\newline 0: \texttt{occluded object}; \texttt{blocked object}; \texttt{covered object}; \texttt{obscured object} \\
 & graspable & peg, gear & 0/1 & 0: \texttt{grasped}; \texttt{held by gripper}; \texttt{picked up}.\newline 1: \texttt{free}; \texttt{loose}; \texttt{unattached}; \texttt{not held} \\
State & gear\_shaft\_exist & circle peg & 1/0 & 1: \texttt{with gear shaft}.\newline 0: \texttt{no gear shaft} \\
 & rotation & hole & 0/1 & 0: \texttt{aligned}; \texttt{axis-aligned orientation}.\newline 1: \texttt{misaligned}; \texttt{rotated in-plane}; \texttt{turned within the image plane}; \texttt{rotated $X$ degrees} ($X\ne0$) \\
 & occupancy & gripper & 1/0 & 1: \texttt{grasping a component}; \texttt{with object between the jaws}.\newline 0: \texttt{grasping nothing}; \texttt{with empty jaws}; \texttt{with space between the jaws} \\
\hline
\end{tabular}
\end{table*}

\subsubsection{Domain Adaptation Fine-Tuning}
Category-agnostic domain adaptation fine-tuning on collected structured annotation data addresses low confidence of pre-trained CLIP models on assembly-specific features.

Training samples one text description per image in each batch and uses a balanced batch sampler to increase exposure to low-frequency physical states such as visibility and rotation. An AdamW optimizer is used: the first stage freezes the vision encoder backbone to stabilize multimodal mappings and mitigate catastrophic forgetting; the second stage unfreezes all parameters for joint fine-tuning.

The standard CLIP contrastive learning loss with a symmetric InfoNCE objective is adopted. The fine-tuned perception network converts low-level continuous visual information into discretized semantic distributions $\{p_{t,i,g}(\ell)\}_{g\in\mathcal{G}}$, using instance-level low-entropy category priors $c_{t,i}$ at the perception frontend. Within-group probability normalization separates the scoring of different physical attributes and quantifies uncertainty over candidate states, bridging raw pixels to downstream symbolic planning and supportive-action reasoning.

\subsection{LLM-Driven Local Task Graph Generation}
Given the discrete instance set $\mathcal{R}_t$ and group-level semantic probability distributions $\{p_{t,i,g}(\ell)\}$, high-level assembly intentions are instantiated as structured local DAGs.

The $M_t$ candidate assembly pairs $p_j=(m_j,f_j,\mathrm{tools}_j)$ are identified from $\mathcal{R}_t$ according to the task objective. Selection relies on geometric compatibility of shape and size, guided by the Mask R-CNN category prior $c_{t,i}$ and RGB-D scale information. Discrete semantic states (mutually exclusive group features such as visibility, graspable, and rotation) and relative occlusion relations organize the output sequence of selected pairs without fixing the final execution order. For each pair $p_j$, a highly abstracted structured world state $w_{t,j}$ is constructed. This state integrates component geometric attributes, discrete states from mutually exclusive semantic groups, end-effector pose, and task goal description. Conditioned on $w_{t,j}$ and the planning history $h_{t,j}$, the LLM autoregressively generates an initial sequence of basic assembly actions $\pi_j\sim P_\phi(\pi\mid w_{t,j},h_{t,j})$, which is parsed into a preliminary chain-like directed graph, as in Algorithm~\ref{alg:main_local_graph}.

A two-stage plan sanity check combined with a history back-injection loop alleviates logical errors and violations of modeled action preconditions in LLM-generated plans:
\begin{itemize}
\item \textbf{Rule-based redundancy elimination}: Domain expert rules automatically remove redundant actions such as consecutive localization of the same object or repeated pick-ups and generate soft guidance (\texttt{GUIDE:}) for refining the plan.
\item \textbf{Dry-run simulation}: State-machine-style precondition verification is performed sequentially for each action in topological order, identifying logical and physical conflicts and generating hard constraints (\texttt{FAIL:}) to avoid.
\end{itemize}

If the preliminary sequence fails the sanity check, a feedback record containing the attempted plan and its \texttt{FAIL:} and \texttt{GUIDE:} messages is added to the planning history $h_{t,j}$ to trigger replanning. This loop permits at most five retry attempts, including the initial attempt, and uses the three most recent feedback records to limit context length, as in Algorithm~\ref{alg:validate_loop}.

\begin{figure}[!t]
\centering
\includegraphics[width=\columnwidth]{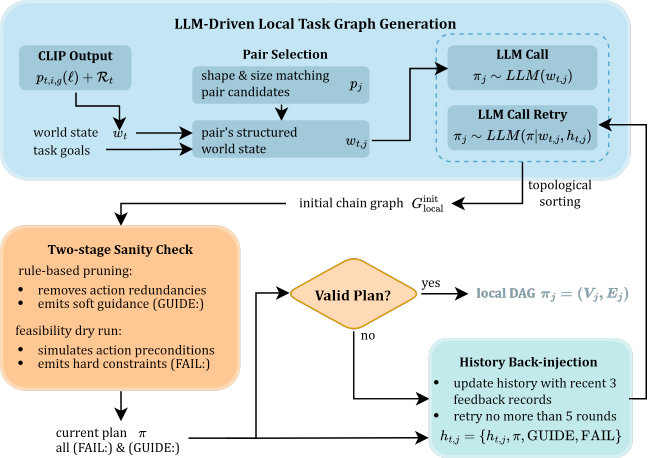}
\caption{LLM local plan generation with two-stage sanity checking and planning history back-injection.}
\label{fig:llm_validation_loop}
\end{figure}

The validated sequence is encapsulated as the local task DAG $\pi_j=(V_j,E_j)$ for downstream supportive action discrimination and global topological coordination. The rules for basic and expanded action vocabularies and the generation workflow are presented in Table~\ref{tab:sanity_rules} and Fig.~\ref{fig:llm_validation_loop}, respectively.

\begin{table*}[!t]
\caption{Two-Stage Plan Sanity Check and History Back-Injection Mechanism for LLM Local Planning}
\label{tab:sanity_rules}
\centering
\renewcommand{\arraystretch}{1.1}
\begin{tabular}{p{0.11\textwidth}p{0.15\textwidth}p{0.38\textwidth}p{0.26\textwidth}}
\hline
\textbf{Stage} & \textbf{Rule / Action} & \textbf{Trigger Condition / Precondition} & \textbf{Output Format} \\
\hline
 & align prune & circular peg or pair already aligned & \texttt{GUIDE:} unnecessary align \\
 & pick prune & male obj. already grasped & \texttt{GUIDE:} avoid repetitive picking \\
Rule-based Layer & localize prune & consecutive \texttt{localize} of the same object & \texttt{GUIDE:} remove redundant localization \\
 & supportive prune & \texttt{release} when gripper is empty; \texttt{remove} when no blocker & \texttt{GUIDE:} supportive action redundancy \\
\hline
 & localize check & Missing/out-of-bounds/inconsistent parameters & \texttt{FAIL:} Invalid parameters \\
 & pick check & Not localized/already grasped/tool conflict & \texttt{FAIL:} Preconditions unmet  \\
 & release check & Male component still grasped/release meaningless & \texttt{FAIL:} Release invalid/wasteful \\
Dry-run Layer & remove check & Gripper not empty/invalid blocker ROI id & \texttt{FAIL:} Remove preconditions unmet \\
 & align check & Incorrect target type/not picked/repetitive alignment & \texttt{FAIL:} Alignment sequence error \\
 & insert check & Not grasped/not aligned/female not localized & \texttt{FAIL:} Insert preconditions unmet \\
Post-check$^{\dagger}$ & remove parallelization & $\ge 2$ consecutive \texttt{remove} & Fan-out/fan-in structural rewiring \\
\hline
Back-Injection & Feedback Set & Any failure occurs & \texttt{tried plan:} + \texttt{FAIL:} + \texttt{GUIDE:} \\
\hline
\end{tabular}
\par\vspace{3pt}
{\scriptsize\raggedright\textit{Note:} $^{\dagger}$Applied only to the expanded-action setting (A2 in Table~\ref{tab:ablation_config}) after a successful dry run. Rewiring is retained only if revalidation succeeds.\par}
\end{table*}

\begin{algorithm}[!t]
\caption{LLM-Driven Local Task Graph Generation}
\label{alg:main_local_graph}
\begin{algorithmic}[1]
\REQUIRE Instance $\mathcal{R}_t$, semantic distributions $\{p_{t,i,g}(\ell)\}_{g\in\mathcal{G}}$, task mode
\ENSURE Set of validated local task DAGs $\Pi$
\STATE $P \gets \mathrm{SelectPairs}(\mathcal{R}_t,\{p_{t,i,g}\},\text{task mode})$
\STATE $\Pi \gets \emptyset$
\FOR{each pair $p_j=(m_j,f_j,\mathrm{tools}_j)\in P$}
    \STATE $w_{t,j} \gets \mathrm{ConstructWorldState}(p_j,\mathcal{R}_t,\{p_{t,i,g}\})$
    \STATE $\pi_j \gets \mathrm{GenerateAndValidate}(w_{t,j},5)$ \COMMENT{Algorithm~\ref{alg:validate_loop}}
    \STATE $\Pi \gets \Pi \cup \{\pi_j\}$ if $\pi_j\ne\mathrm{failure}$
\ENDFOR
\RETURN $\Pi$
\end{algorithmic}
\end{algorithm}

\begin{algorithm}[!t]
\caption{Validation and Planning History Back-Injection}
\label{alg:validate_loop}
\begin{algorithmic}[1]
\REQUIRE World state $w_{t,j}$, retry-attempt limit $K$
\ENSURE Validated local DAG $\pi_j=(V_j,E_j)$ or $\mathrm{failure}$
\STATE $h_{t,j} \gets \emptyset$ \COMMENT{Planning history}
\FOR{$k=1$ to $K$}
    \STATE $\pi \gets \mathrm{LLM}(w_{t,j},h_{t,j})$ 
    \STATE $G' \gets \mathrm{BuildInitialChainGraph}(\pi)$
    \STATE $(G',\mathrm{GUIDE}) \gets \mathrm{RulePrune}(G',w_{t,j})$
    \STATE $(\mathrm{valid},\mathrm{FAIL}) \gets \mathrm{DryRun}(G',w_{t,j})$
    \IF{$\mathrm{valid}$}
        \RETURN $G'$
    \ENDIF
    \STATE $h_{t,j} \gets \mathrm{UpdateHistory}(h_{t,j},\pi,\mathrm{GUIDE},\mathrm{FAIL})$ 
\ENDFOR
\RETURN $\mathrm{failure}$
\end{algorithmic}
\end{algorithm}

\subsection{JS-Aug: Divergence-Driven Supportive Action Reasoning and Local Graph Augmentation}
\begin{figure}[!t]
\centering
\subfloat[]{\includegraphics[width=0.3\columnwidth]{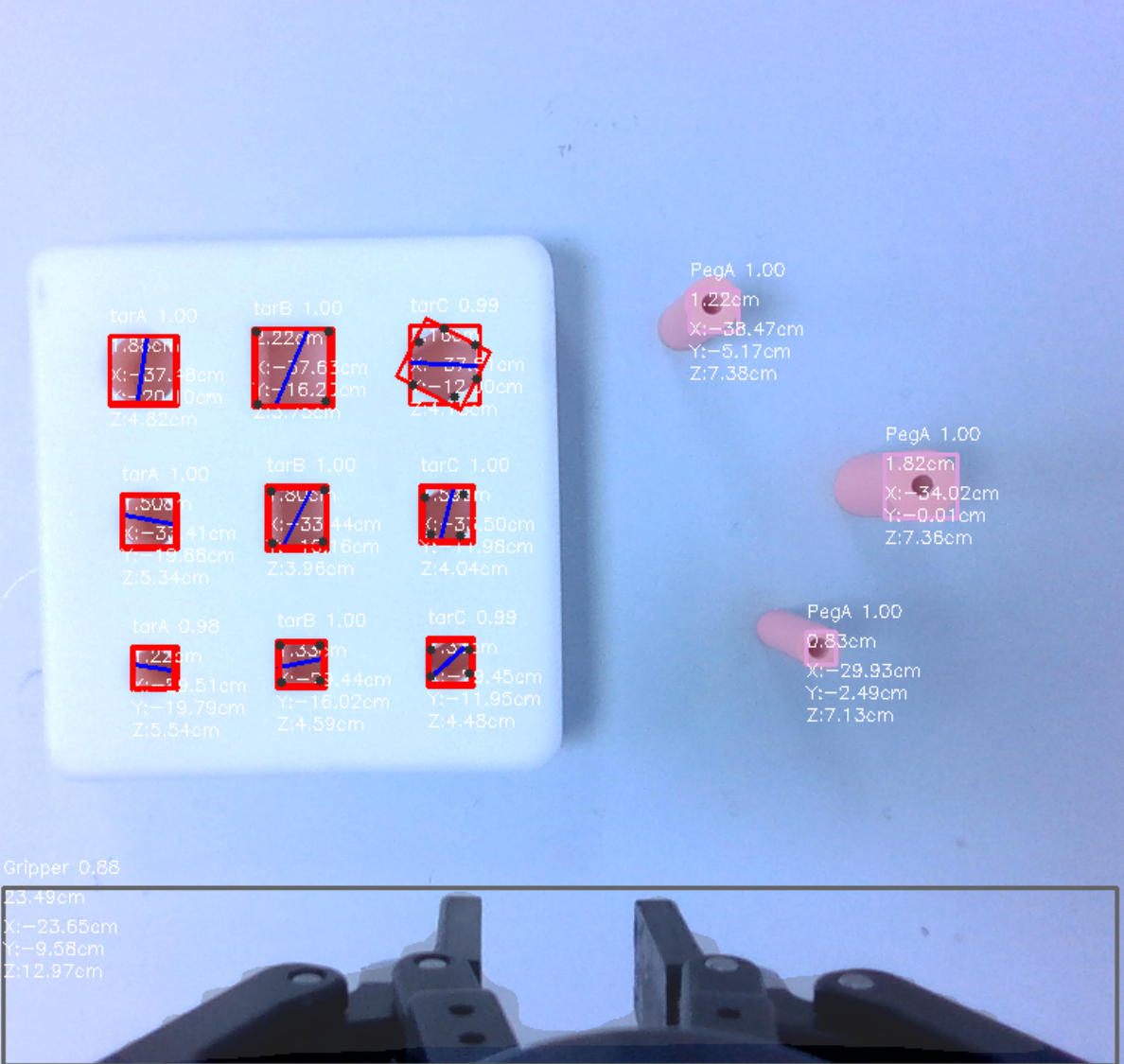}}\hfill
\subfloat[]{\includegraphics[trim=300bp 0bp 220bp 0bp,clip,width=0.3\columnwidth]{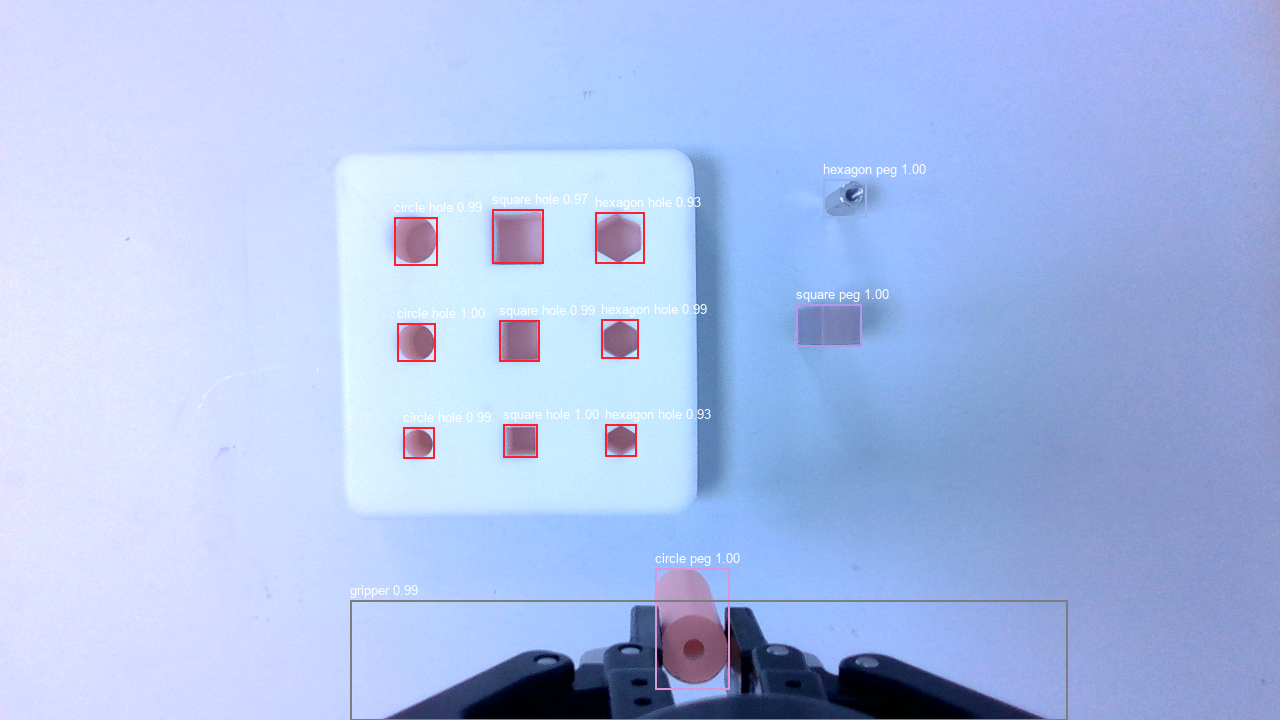}}\hfill
\subfloat[]{\includegraphics[width=0.3\columnwidth]{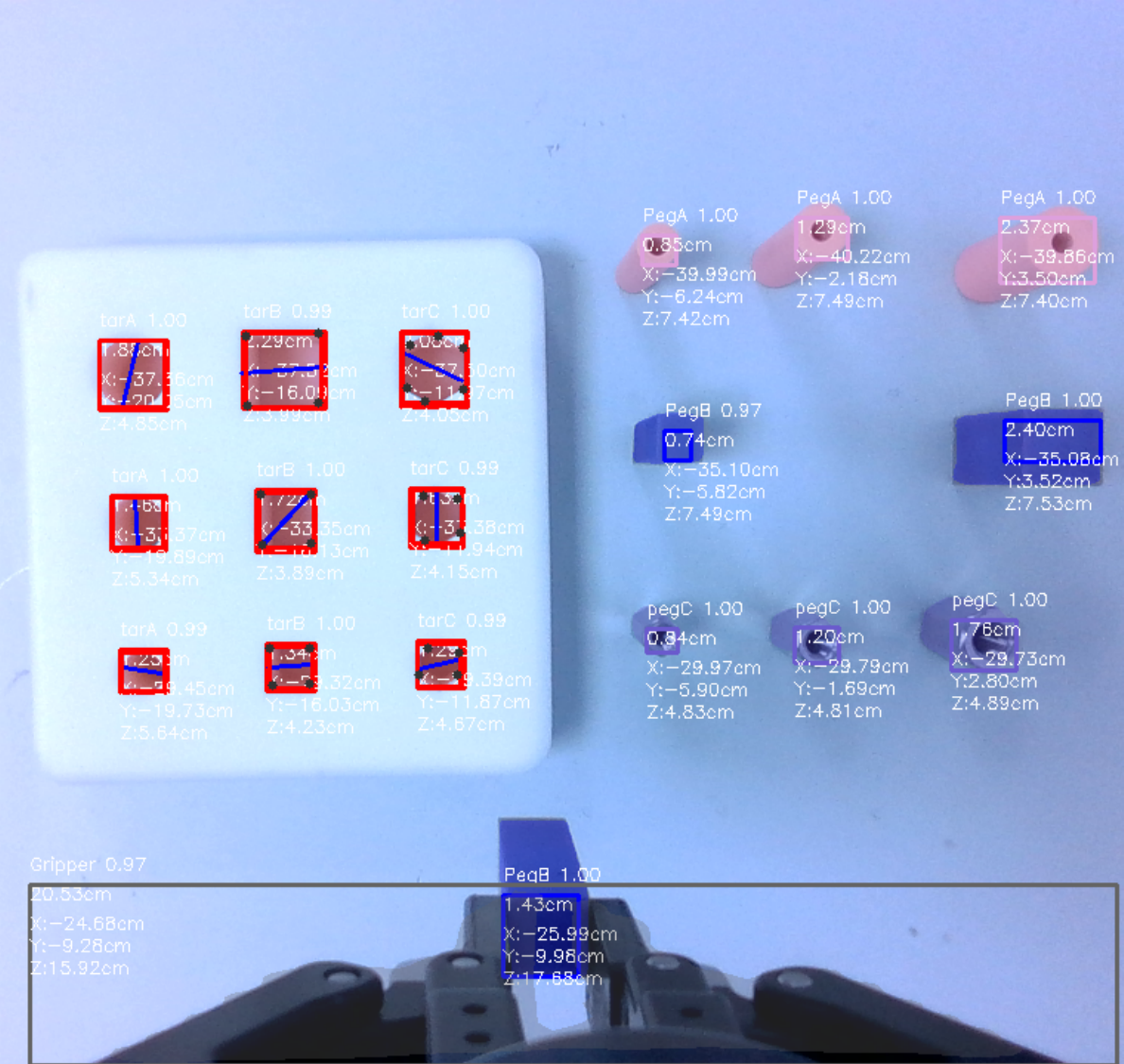}}
\\
\subfloat[]{\includegraphics[width=0.3\columnwidth]{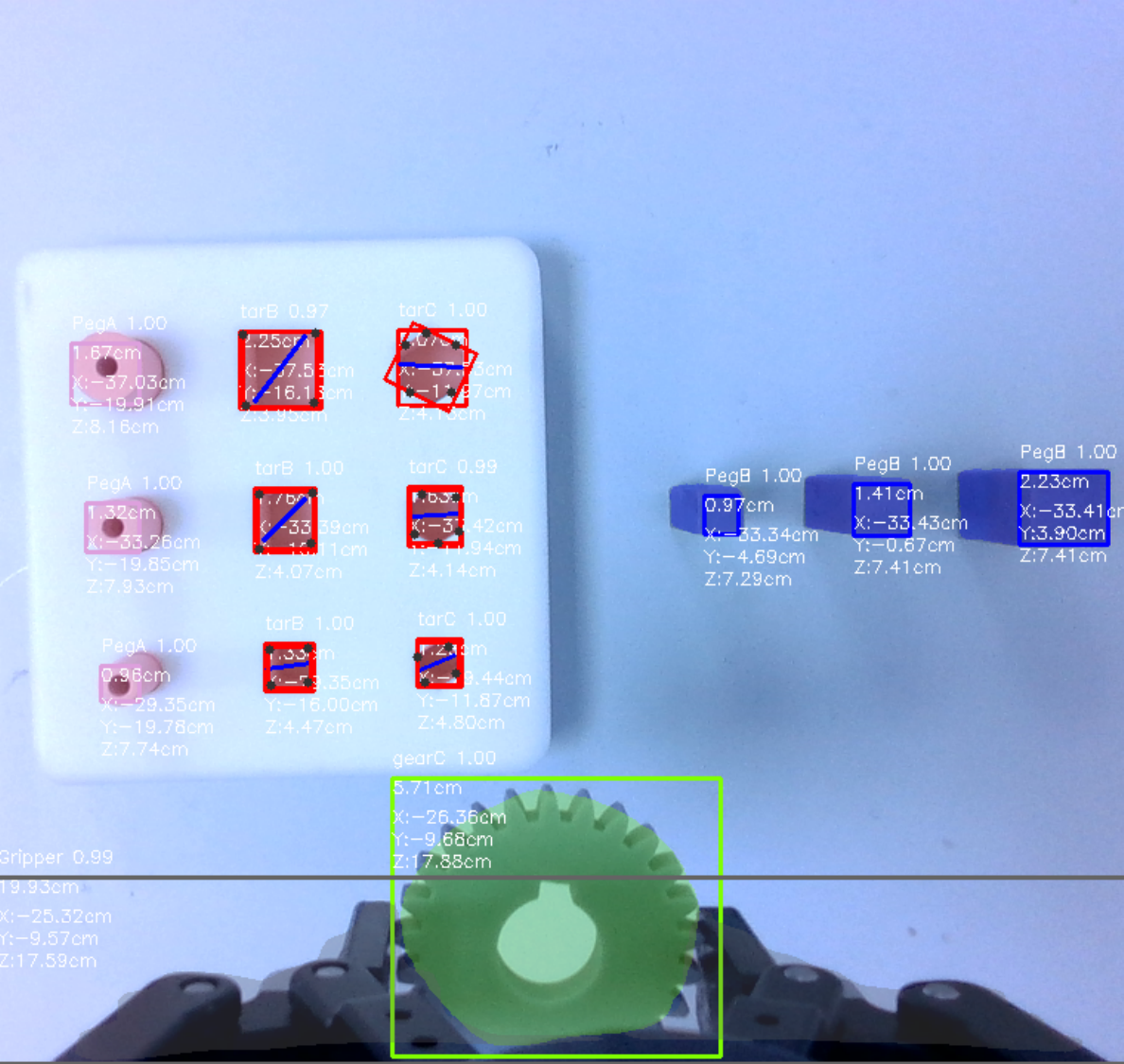}}\hfill
\subfloat[]{\includegraphics[width=0.3\columnwidth]{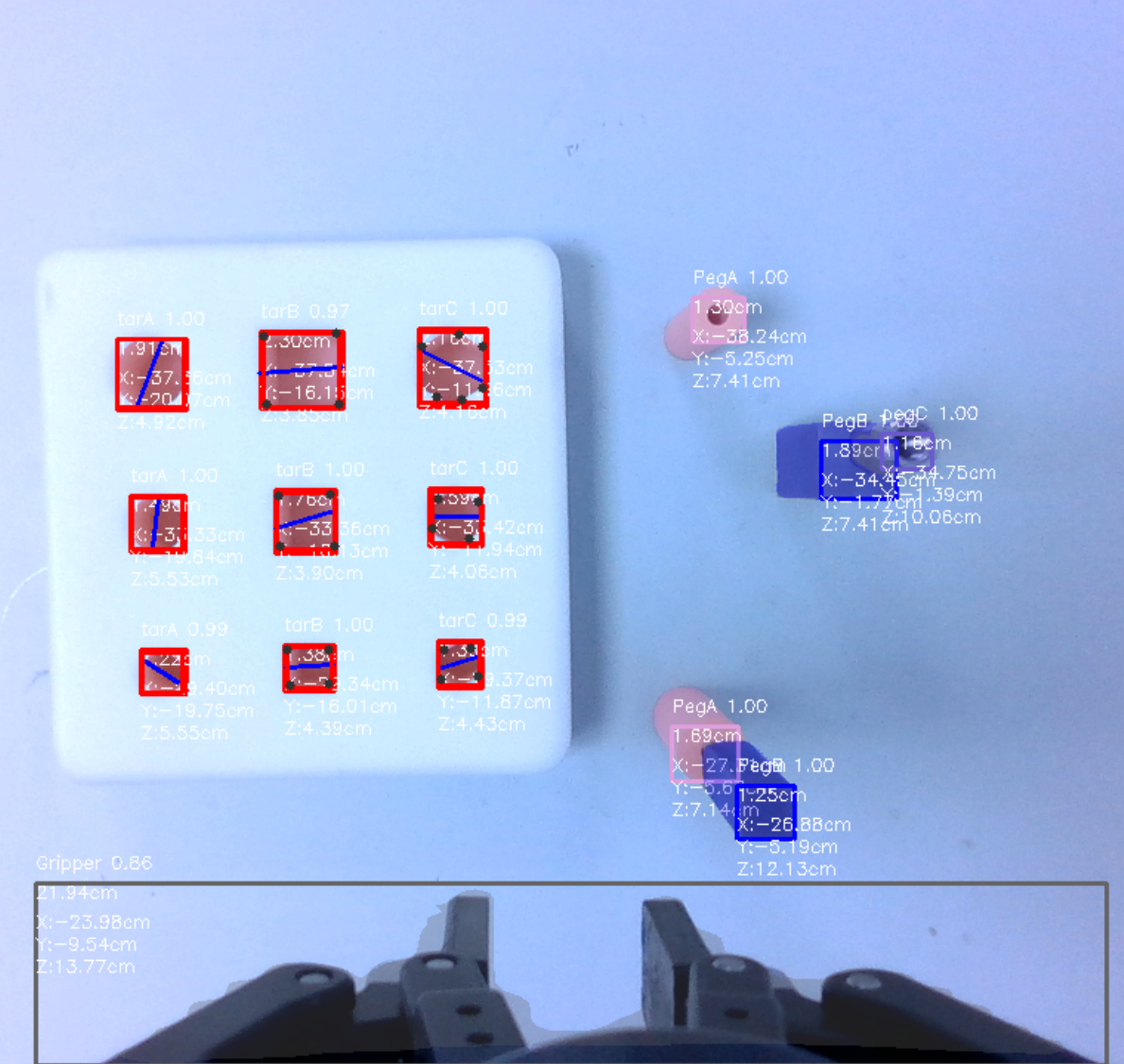}}\hfill
\subfloat[]{\includegraphics[width=0.3\columnwidth]{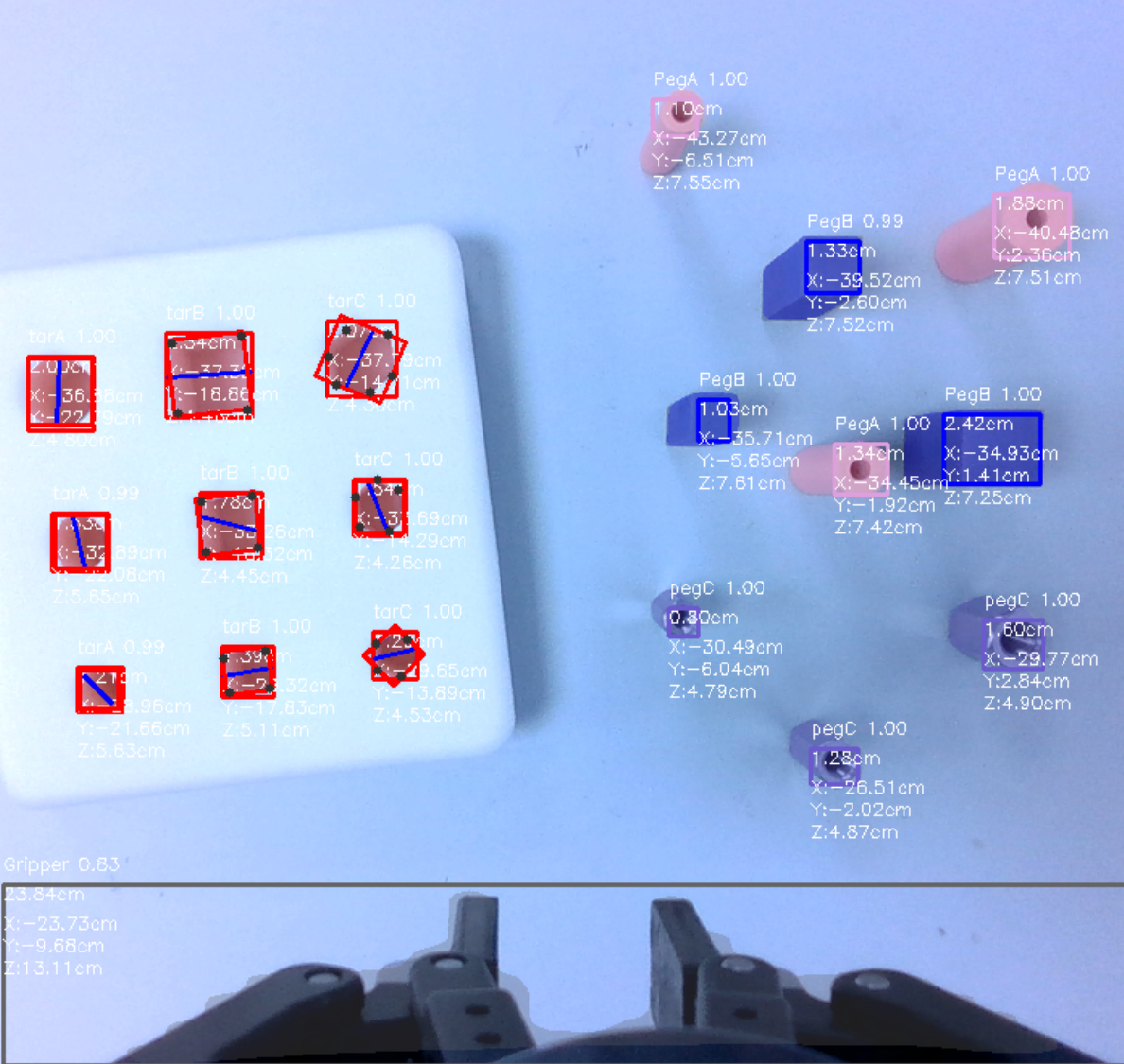}}
\caption{Canonical reference and variants: (a) standard independent circular peg-in-hole; (b) circular peg-in-hole with the male target grasped; (c) non-circular peg-in-hole; (d) occupancy requiring \texttt{release}; (e) stacking requiring \texttt{remove}; (f) misalignment requiring \texttt{align}. No supportive actions are required in (a)--(c).}
\label{fig:base_var}
\end{figure}

Operation sequences in complex multi-pair assembly tasks exhibit strong structural commonality. The framework decouples assembly logic into basic actions and supportive actions. The LLM generates local DAGs $\pi_j$ composed solely of basic actions \texttt{localize}, \texttt{pick}, \texttt{insert} from structured world states. Supportive actions that resolve environmental deviations—such as occlusion clearance, occupancy release, and pose alignment—are decoupled from LLM planning and handled by the JS-Aug module. Representative baseline and variant scenes appear in Fig.~\ref{fig:base_var}.

\begin{figure}[tp]
\centering
\includegraphics[width=\columnwidth]{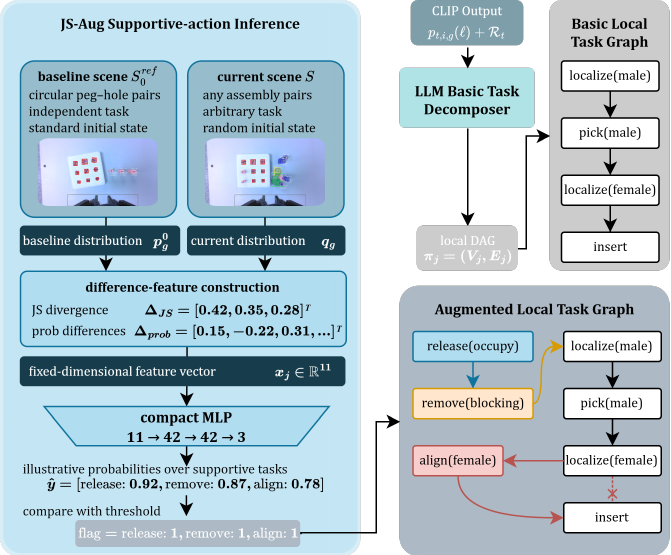}
\caption{JS-Aug pipeline: difference features constructed from baseline and current distributions, supportive action trigger flags predicted, and structured augmentation performed on the local DAG.}
\label{fig:jstrans_overview}
\end{figure}

One independent circular peg-in-hole assembly under standard initial conditions serves as the canonical reference $S_0^{\mathrm{ref}}$. It requires no supportive actions due to absent physical occlusions, gripper occupancy, and rotational misalignment, providing a stable reference for quantifying scene deviations. JS-Aug compares semantic probabilities against this reference using a compact MLP. The overall input-output pathway and graph augmentation logic are illustrated in Fig.~\ref{fig:jstrans_overview}.

\subsubsection{Difference Feature Construction}
JS-Aug constructs baseline-variant difference features directly at the mutually exclusive semantic group level without additional text fine-tuning. For each candidate pair $j$, given the current scene distributions $\{q_g\}$ and baseline distributions $\{p_g^0\}$, L1 normalization is applied, with $\varepsilon>0$ for numerical stability:
\begin{equation}
p_g^0 \leftarrow \frac{p_g^0}{\sum_i p_{g,i}^0 + \varepsilon},\qquad
q_g \leftarrow \frac{q_g}{\sum_i q_{g,i} + \varepsilon}.
\end{equation}
The JSD between the distributions is computed:
\begin{equation}
\mathrm{JS}(p_g^0,q_g)=\tfrac12\mathrm{KL}(p_g^0\|m_g)+\tfrac12\mathrm{KL}(q_g\|m_g),
\end{equation}
where $\mathrm{KL}(a\|b)=\sum_i a_i\log_2\tfrac{a_i}{b_i}$, $m_g=\tfrac12(p_g^0+q_g)$.

Mutually exclusive groups are aggregated by component role:
\begin{equation}
\begin{aligned}
\text{male}  &\colon \{\texttt{visibility}, \texttt{graspable}\}, \\
\text{female} &\colon \{\texttt{rotation}\}, \\
\text{tool} &\colon \{\texttt{occupancy}\}.
\end{aligned}
\end{equation}
Averaging the JS divergence within each role yields a 3-D summary vector $\Delta_{\mathrm{JS}}=[\overline{\mathrm{JS}}_{\text{male}},\overline{\mathrm{JS}}_{\text{female}},\overline{\mathrm{JS}}_{\text{tool}}]$. Eight signed group-probability differences $\Delta_{\mathrm{prob}}=\mathrm{concat}_g(q_g-p_g^0)$ are retained. Baseline and current distributions share the same intra-group label order. Their concatenation gives $x_j=[\Delta_{\mathrm{JS}}\Vert\Delta_{\mathrm{prob}}]\in\mathbb{R}^{11}$.

\subsubsection{MLP Reasoning and Structured Graph Augmentation}
The vector $x_j$ is fed to a compact MLP with two 42-unit ReLU hidden layers and sigmoid outputs to predict supportive-action trigger probabilities $\hat y\in[0,1]^{|\mathcal{A}_{\mathrm{supp}}|}$. Independent decision thresholds binarize the outputs to produce trigger flags $\mathrm{flag}_k$. After applicability checks, active flags trigger the corresponding graph edits, as in Algorithm~\ref{alg:jstrans}.

Table~\ref{tab:jstrans_rules} defines graph edits around $v_{\mathrm{male}}$ and $v_{\mathrm{female}}$, the \texttt{localize} nodes of the male and female components. These edits use the corresponding ROI associations and blocker relations. After topological augmentation, self-loops, duplicate edges, and edges with invalid endpoints are removed from the edge set. For nodes resolvable to valid ROIs, parameter binding updates node parameters with geometric information (target position and orientation) associated with the current scene to align the local plan graph with low-level physical observations.

\begin{table}[!t]
\caption{Graph Augmentation Rules for Supportive Actions}
\label{tab:jstrans_rules}
\centering
\renewcommand{\arraystretch}{1.1}
\setlength{\tabcolsep}{3.5pt}
\begin{tabular}{p{0.14\columnwidth}p{0.80\columnwidth}}
\hline
\textbf{Action} & \textbf{Graph Editing} \\
\hline
\texttt{align} & Insert $v_{\mathrm{align}}$ after $v_{\mathrm{female}}$; replace all $(v_{\mathrm{female}},u)$ with $(v_{\mathrm{align}},u)$ and append $(v_{\mathrm{female}},v_{\mathrm{align}})$ \\
\texttt{remove} & Insert $v_{\mathrm{rem},i}$ for each blocker $i$ and append $(v_{\mathrm{rem},i},v_{\mathrm{male}})$; keep \texttt{remove} nodes independent \\
\texttt{release} & Insert $v_{\mathrm{rel}}$ with edges $(v_{\mathrm{rel}},r)$ to all pre-existing roots $r$ \\
\hline
\end{tabular}
\end{table}

\begin{algorithm}[!t]
\caption{Divergence-Driven Local Task Augmentation}
\label{alg:jstrans}
\begin{algorithmic}[1]
\REQUIRE Local DAG $\pi_j$, current distributions $\{q_g\}$, baseline distributions $\{p_g^0\}$
\ENSURE Augmented local DAG $\pi'_j$
\STATE Normalization: $q_g \gets q_g/(\sum q_g+\varepsilon)$; $p_g^0 \gets p_g^0/(\sum p_g^0+\varepsilon)$
\FOR{each group $g$}
    \STATE Compute $\mathrm{JS}(p_g^0,q_g)$ (base 2)
\ENDFOR
\STATE $\Delta_{\mathrm{JS}} \gets$ average JS divergence by component role
\STATE $\Delta_{\mathrm{prob}} \gets \mathrm{concat}_g(q_g-p_g^0)$ (fixed key order)
\STATE $x_j \gets [\Delta_{\mathrm{JS}}\Vert\Delta_{\mathrm{prob}}]\in\mathbb{R}^{11}$
\STATE $\hat y \gets \mathrm{sigmoid}(\mathrm{MLP}_{\theta}(x_j))$
\FOR{$k\in\mathcal{A}_{\mathrm{supp}}$}
    \STATE $b_k \gets \mathbf{1}\{\hat y_k>\tau_k\}$; $\mathrm{flag}_k\gets b_k$
\ENDFOR
\IF{circular peg-in-hole or gear assembly task}
    \STATE $\mathrm{flag}_{\mathrm{align}}\gets 0$
\ENDIF
\STATE Locate anchor nodes $v_{\mathrm{male}},v_{\mathrm{female}}$ in $\pi_j$
\STATE $\pi'_j \gets \mathrm{AugmentGraph}(\pi_j,\{\mathrm{flag}_k\})$ \COMMENT{see Table~\ref{tab:jstrans_rules}}
\STATE Sanitize edges; bind parameters with current geometric information
\RETURN $\pi'_j$
\end{algorithmic}
\end{algorithm}

\subsection{TOPO: Token Precondition/Effect Projection and Global Topological Coordination}
\begin{figure*}[!t]
\centering
\includegraphics[width=\textwidth]{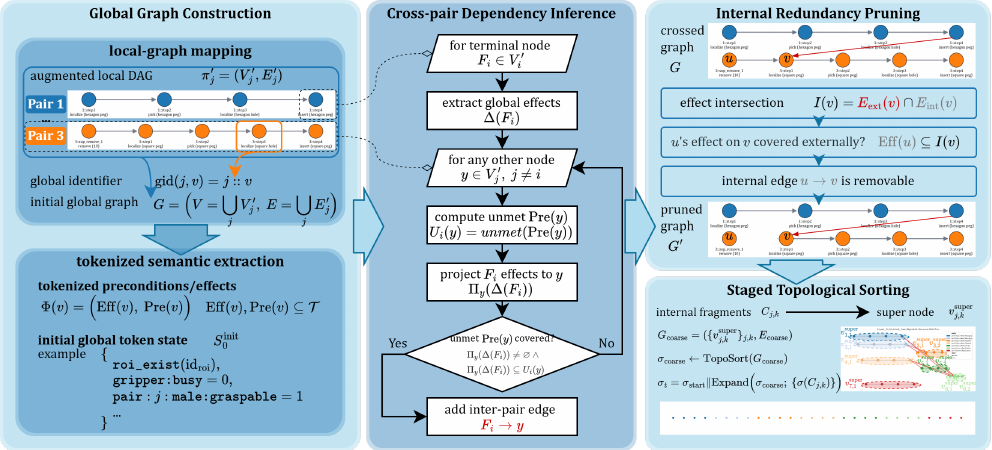}
\caption{TOPO overview: tokenized semantic extraction, cross-pair dependency inference, internal redundancy pruning, and staged topological sorting.}
\label{fig:topo_overview}
\end{figure*}

\begin{figure}[!t]
\centering
\includegraphics[width=\columnwidth]{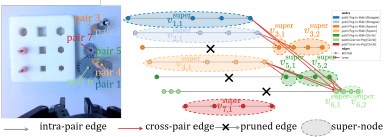}
\caption{Global topological coordination example.}
\label{fig:topo_pruning_example}
\end{figure}

The set of augmented local task graphs $\{\pi'_j\}_{j=1}^{M_t}$ is coordinated into a single global sequence $\sigma_t$ for downstream execution through the TOPO module. The module abstracts dependency reasoning from symbolic planning into a token-algebraic coordination heuristic. It follows a pipeline of global graph construction, cross-pair dependency inference, internal redundancy pruning, and staged topological sorting. This mitigates global topological conflicts that may arise from the LLM. The overall workflow and visualization examples are shown in Fig.~\ref{fig:topo_overview} and Fig.~\ref{fig:topo_pruning_example}.

\subsubsection{Global Directed Graph Construction}
Nodes in each augmented local task graph $\pi'_j=(V'_j,E'_j)$ are mapped to globally unique identifiers to eliminate namespace collisions among pairs:
\begin{equation}
\mathrm{gid}(j,v)=j\!:\!:\!v,\qquad j=1,\ldots,M_t,\ v\in V'_j.
\end{equation}
Then, the initial global directed graph is constructed:
\begin{equation}
G=\bigl(V=\bigcup_j V'_j,\ E=\bigcup_j E'_j\bigr),
\end{equation}
where the initial edge set $E$ preserves all intra-pair dependencies, with local graphs without explicit topological edges defaulting to implicit chain dependencies.

\subsubsection{Tokenized Semantic Extraction and Cross-Pair Dependency Inference}
A discrete global token universe $\mathcal{T}$ includes atomic propositions such as ROI existence, gripper state, and pair-local semantics. For any global node $v \in V$, precondition and effect sets are extracted:
\begin{equation}
\Phi(v)=\bigl(\mathrm{Eff}(v),\ \mathrm{Pre}(v)\bigr),\qquad \mathrm{Eff}(v),\mathrm{Pre}(v)\subseteq\mathcal{T}.
\end{equation}
TOPO selects the starting pair by choosing a shortest local sequence whose root preconditions hold in the initial scene state. Provider $F_i$ and its consumers use the scene state $S_i^{\mathrm{init}}$, updated by the starting pair's terminal global effects only for the remaining providers. Pair-local states remain unchanged. The unmet preconditions for node $x$ are
\begin{equation}
U_i(x)=\{t\in\mathrm{Pre}(x)\mid S_i^{\mathrm{init}}\not\models t\}.
\end{equation}
Cross-pair dependency inference relies on algebraic projection, as in Algorithm~\ref{alg:topo_cross_pair}. The terminal action node $F_i$ (typically \texttt{insert}) of pair $i$ serves as the dependency provider. Its deterministic net change to the global state is extracted:
\begin{equation}
\Delta(F_i)\subseteq\mathcal{T}_{\mathrm{global}},
\end{equation}
where no-operation effects relative to $S_i^{\mathrm{init}}$ are pre-filtered. For any node $y\in V'_j$ ($j\neq i$), the cross-pair effect projection subset is:
\begin{equation}
\Pi_y(\Delta(F_i))=\{t\in\Delta(F_i)\mid \mathrm{key}(t)\in\mathrm{Keys}(\mathrm{Pre}(y))\}.
\end{equation}
A cross-pair directed edge $F_i\to y$ is added to $G$ if and only if the projection is a nonempty subset of the unmet preconditions of $y$:
\begin{equation}
\Pi_y(\Delta(F_i))\neq\varnothing\ \wedge\ \Pi_y(\Delta(F_i))\subseteq U_i(y).
\end{equation}

For multi-solution states, two deterministic safeguards are applied: (i) when $M_t>1$ and no cross-pair edges are inferred, minimal sequential constraints (e.g., $F_{i_k}\to\mathrm{Root}(i_{k+1})$) are added according to input order to prevent interleaving of unrelated pairs; (ii) disconnected components of the coarse graph are sorted separately and concatenated in input order.

\subsubsection{Effect Intersection and Redundancy Pruning}
Predecessors of $v$ are partitioned into external and internal sets, aggregating only effects that change the initial scene state:
\begin{equation}
\begin{split}
E_{\mathrm{ext}}(v)=&\bigcup_{u\in\mathrm{ExtPred}(v)}\mathrm{Eff}(u),\\
E_{\mathrm{int}}(v)=&\bigcup_{u\in\mathrm{IntPred}(v)}\mathrm{Eff}(u).
\end{split}
\end{equation}
The effect intersection is defined as $I(v)=E_{\mathrm{ext}}(v)\cap E_{\mathrm{int}}(v)$. When $I(v)\neq\varnothing$, an internal directed edge $u\to v$ with $\mathrm{Eff}(u)\subseteq I(v)$ is removed, as its state contribution to $v$ is fully covered by external constraints. Redundant branches are omitted from the output, with bypass edges preserving local reachability where needed, as in Algorithm~\ref{alg:topo_pruning}.

\subsubsection{Staged Topological Sorting}
A conventional node-level topological sort respects graph dependencies but can interleave actions from different pairs. Staged topological sorting keeps internal action fragments contiguous, limiting gripper conflicts when an initially satisfied empty-gripper condition is absent from the projection, as in Algorithm~\ref{alg:topo_staged}.

For each pair $j$, internal edges are cut at nodes incident to cross-pair edges in $G_{\mathrm{int},j}$. The resulting weakly connected components form non-overlapping internal action fragments:
\begin{equation}
\mathcal{C}_j = \{C_{j,k}\}_{k=1}^{K_j},
\end{equation}
where each fragment $C_{j,k}$ is topologically ordered internally. Each fragment is contracted into a global virtual super-node $v_{j,k}^{\mathrm{super}}$, forming a coarse-grained graph:
\begin{equation}
G_{\mathrm{coarse}} = \bigl(\{v_{j,k}^{\mathrm{super}}\}_{j,k},\ E_{\mathrm{coarse}}\bigr).
\end{equation}
Topological sorting is performed on the coarse-grained graph:
\begin{equation}
\sigma_{\mathrm{coarse}} \gets \mathrm{TopoSort}(G_{\mathrm{coarse}}).
\end{equation}
The starting sequence $\sigma_{\mathrm{start}}$ precedes the expanded fragments, with its nodes omitted during expansion:
\begin{equation}
\sigma_t = \sigma_{\mathrm{start}}\Vert\mathrm{Expand}\bigl(\sigma_{\mathrm{coarse}};\ \{\sigma(C_{j,k})\}\bigr).
\end{equation}
This packing strategy preserves the represented dependencies while keeping each fragment contiguous.

\begin{algorithm}[!t]
\caption{Cross-Pair Dependency Inference}
\label{alg:topo_cross_pair}
\begin{algorithmic}[1]
\REQUIRE Global graph $G=(V,E)$, starting states $\{S_i^{\mathrm{init}}\}$, terminal node set $\{F_i\}$
\ENSURE Cross-pair edge set $E_{\mathrm{cross}}$
\STATE $E_{\mathrm{cross}} \gets \emptyset$
\FOR{each pair $i$}
    \STATE $\Delta(F_i) \gets \mathrm{ExtractGlobalEffects}(F_i,S_i^{\mathrm{init}})$ 
    \FOR{each $j\neq i$, each node $y\in V'_j$}
        \STATE $U_i(y) \gets \{t\in\mathrm{Pre}(y)\mid S_i^{\mathrm{init}}\not\models t\}$
        \STATE $\Pi_y(\Delta(F_i)) \gets \{t\in\Delta(F_i)\mid \mathrm{key}(t)\in\mathrm{Keys}(\mathrm{Pre}(y))\}$
        \IF{$\Pi_y(\Delta(F_i))\neq\varnothing$ \textbf{and} $\Pi_y(\Delta(F_i))\subseteq U_i(y)$}
            \STATE $E_{\mathrm{cross}} \gets E_{\mathrm{cross}}\cup\{F_i\to y\}$
        \ENDIF
    \ENDFOR
\ENDFOR
\RETURN $E_{\mathrm{cross}}$
\end{algorithmic}
\end{algorithm}

\begin{algorithm}[!t]
\caption{Internal Redundancy Pruning}
\label{alg:topo_pruning}
\begin{algorithmic}[1]
\REQUIRE Global graph $G=(V,E)$
\ENSURE Pruned graph $G'$
\STATE $G'\gets G$
\FOR{each node $v\in V$}
    \STATE $E_{\mathrm{ext}}(v) \gets \bigcup_{u\in\mathrm{ExtPred}(v)}\mathrm{Eff}(u)$
    \STATE $E_{\mathrm{int}}(v) \gets \bigcup_{u\in\mathrm{IntPred}(v)}\mathrm{Eff}(u)$
    \STATE $I(v) \gets E_{\mathrm{ext}}(v)\cap E_{\mathrm{int}}(v)$
    \FOR{each internal predecessor $u$ of $v$}
        \IF{$I(v)\neq\varnothing$ \textbf{and} $\mathrm{Eff}(u)\subseteq I(v)$}
            \STATE remove edge $u\to v$ from $G'$
        \ENDIF
    \ENDFOR
\ENDFOR
\STATE exclude redundant branches and reconnect retained nodes
\RETURN $G'$
\end{algorithmic}
\end{algorithm}

\begin{algorithm}[!t]
\caption{Staged Topological Sorting}
\label{alg:topo_staged}
\begin{algorithmic}[1]
\REQUIRE Global graph $G=(V,E)$, starting sequence $\sigma_{\mathrm{start}}$
\ENSURE Global sequence $\sigma_t$
\IF{$G$ contains a directed cycle}
    \RETURN failure
\ENDIF
\FOR{each pair $j$}
    \STATE cut internal edges at cross-pair contacts to form $\mathcal{C}_j$
    \STATE contract each fragment $C_{j,k}$ into super-node $v_{j,k}^{\mathrm{super}}$
\ENDFOR
\STATE construct coarse-grained graph $G_{\mathrm{coarse}}$
\IF{$G_{\mathrm{coarse}}$ contains a directed cycle}
    \RETURN failure
\ENDIF
\STATE $\sigma_{\mathrm{coarse}} \gets \mathrm{TopoSort}(G_{\mathrm{coarse}})$
\STATE $\sigma_t\gets\sigma_{\mathrm{start}}\Vert\mathrm{Expand}(\sigma_{\mathrm{coarse}};\{\sigma(C_{j,k})\})$
\RETURN $\sigma_t$
\end{algorithmic}
\end{algorithm}

\section{BT Management and Force-Aware Execution}
The BT management and force-aware execution module of the neuro-symbolic framework receives the globally coordinated discrete sequence $\sigma_t$ from the TOPO module and compiles it at runtime into an interpretable behavior tree. The BT asynchronously schedules low-level atomic action skills while monitoring 6-D Force/Torque (F/T) signals in real time. Each action node includes self-check and fine-tuning cycles to form a local feedback loop between the symbolic action and its physical execution. Detected contact anomalies or skill failures trigger safety interruption and recovery for long-horizon assembly.

\subsection{BT-Based Skill Scheduling and Local Self-Check Mechanisms}
Given the global sequence $\sigma_t$, the BT manager normalizes each step through a structured interface (action type, target ROI identifier, geometric and pose parameters) to ensure consistency between high-level semantics and low-level controllers. The sequence is mapped to a behavior tree, with discrete steps instantiated as atomic action nodes that adhere to standard BT return semantics.%

Each action node encapsulates closed-loop verification and local correction:
\begin{itemize}
\item \texttt{localize}: The end-effector is coarsely positioned to a predefined safe height, then enters a high-precision visual servoing loop driven by the Eye-on-Hand camera. The system iteratively adjusts the pose to center the target feature in the gripper base coordinate frame.
\item \texttt{pick}: After execution, the system parses low-level gripper register feedback (e.g., motor current thresholds and \texttt{gOBJ} state bit from Robotiq) for rule-based grasp success verification. If current is too low or \texttt{gOBJ} indicates no object, the node suspends and triggers up to three local retries. After successful pick, drop-guard monitoring is activated; a sudden current drop or \texttt{gOBJ} state flip returns \texttt{FAILURE}, prompting the BT root to initiate recovery.
\item \texttt{insert}: The system invokes a dedicated high-precision insertion skill node that employs a hybrid force-control strategy combining Archimedean spiral blind search and Proximal Policy Optimization (PPO) RL impedance control.
\item \texttt{release}: Geometric collision constraints and semantic cues from the frontend CLIP are used to compute a safe placement zone. The occupied object is transferred smoothly to this zone and released, clearing spatial interference and restoring gripper availability.
\item \texttt{remove}: This action reuses the \texttt{localize} $\to$ \texttt{pick} $\to$ \texttt{release} behavioral subtree. After localizing the blocker ROI, the object is grasped, transported to a safe isolation area, and released.
\item \texttt{align}: The minimal equivalent correction angle is computed from the geometric symmetry of the target component. Circular targets require no alignment due to rotational invariance.
\end{itemize}

\subsection{Hybrid Insertion Skill via Spiral Search and RL Impedance Control}
For the precision insertion phase, a dedicated hybrid force-control skill server combines Archimedean spiral blind search for reducing macroscopic pose errors with a PPO RL impedance control policy for fine-tuning during the contact-rich phase.

\subsubsection{Safe Approach} The end-effector approaches along the designated insertion axis at constant downward velocity while performing coarse XY-plane pose correction in free space.

\subsubsection{Contact Blind Search} Upon detection of initial rigid contact by the Robotiq F/T sensor, a constant-force downward Archimedean spiral trajectory search of maximum radius $10\,\mathrm{mm}$ originates from the contact point to locate the chamfer entrance. A significant depth drop along the Z-axis relative to the initial contact plane indicates successful capture of the hole feature.

\subsubsection{Fine-Tuning Insertion} Once inside the restricted hole space, control authority transitions to the PPO policy. The policy outputs a fine-tuning vector $\mathbf{a}_t\in[-1,1]^7$ in a continuous 7-dimensional action space. Small planar displacements and axis-angle rotations are mapped as:
\begin{equation}
\begin{split}
\Delta x = 0.001\, a_t[0],\quad \Delta y = 0.001\, a_t[1],\\
\theta = a_t[3]\cdot\frac{\pi}{180},\quad \boldsymbol{\omega}=\theta\,\mathbf{u},\ \mathbf{u}=a_t[4{:}6].
\end{split}
\end{equation}
Pose updates employ exponential mapping and composition in SO(3):
\begin{equation}
R_{t+1} = R_t \circ \exp([\boldsymbol{\omega}]_\times).
\end{equation}

The matrix $[\boldsymbol{\omega}]_\times$ is the skew-symmetric matrix associated with $\boldsymbol{\omega}$. The RL policy employs two mechanisms: 
\begin{itemize}
\item \textbf{Multi-Frame Observation Fusion:} Up to 20 control cycles of raw 6-D wrench vectors $\mathbf{w}=[f_x,f_y,f_z,\tau_x,\tau_y,\tau_z]=[\boldsymbol{f},\boldsymbol{\tau}]$ are sampled per RL decision step. The moving average $\overline{\mathbf{w}}$ and absolute instantaneous peak $\mathbf{w}^{\max}$ form a low-noise 12-D observation:
\begin{equation}
\mathbf{o}_t = [\overline{\mathbf{w}};\ \mathbf{w}^{\max}]\in\mathbb{R}^{12}.
\end{equation}

\item \textbf{Reward Function Shaping:} Policy inputs are restricted to F/T proprioceptive features to enable zero-shot sim-to-real transfer. During training, the reward function uses privileged simulation states for shaping:
\begin{equation}
\label{eq:ppo_reward}
\begin{split}
r &= w_1\,\Delta z_{\mathrm{insert}} + w_2\,K_{xy} \\
&- w_3\bigl(\beta_1\|\overline{\boldsymbol{f}}\|_2^2 + \beta_2\|\overline{\boldsymbol{\tau}}\|_2^2 + \eta\|\mathbf{w}^{\max}\|_{\infty}\bigr)\\
&+ w_4\,\mathbb{I}(|p_z - p^{\mathrm{hole}}_z|<\epsilon),
\end{split}
\end{equation}
where $w_*$ balances each term, $\Delta z_{\mathrm{insert}} = z_{t-1}-z_t$ is the actual insertion depth increment, $K_{xy}=\exp(-\alpha\|p_{xy}-p^{\mathrm{hole}}_{xy}\|_2^2)$ rewards XY-plane concentric alignment, the penalty term suppresses excessive average forces and destructive instantaneous peaks, and the final term provides a sparse task-completion reward.
\end{itemize}

\subsection{F/T Monitoring and Global Anomaly Recovery}
The BT manager continuously subscribes to the low-level force-control data stream throughout the control cycle. Execution is interrupted and the exception-handling branch is activated when any downstream action node returns \texttt{FAILURE} or the real-time wrench signal exceeds preset safety contact thresholds.

Upon triggering recovery, a retreat procedure is executed: vertical lift along the Z-axis to a safe withdrawal plane, retraction to a predefined retreat observation pose, gripper state reset, and a limited number of attempts to re-enter the task sequence. The hybrid insertion skill incorporates millisecond-level hard-interrupt monitoring of F/T peaks. On irreversible contact collisions, the skill aborts, propagates the \texttt{FAILURE} flag upward through the tree, and enables unified safety scheduling from high-level cognition to low-level physical hardware.

\section{Experimental Setup and Evaluation Protocol}
\subsection{Data and Platforms}
\label{sec:setup_data}
\begin{figure}[!t]
\centering
\includegraphics[width=\columnwidth]{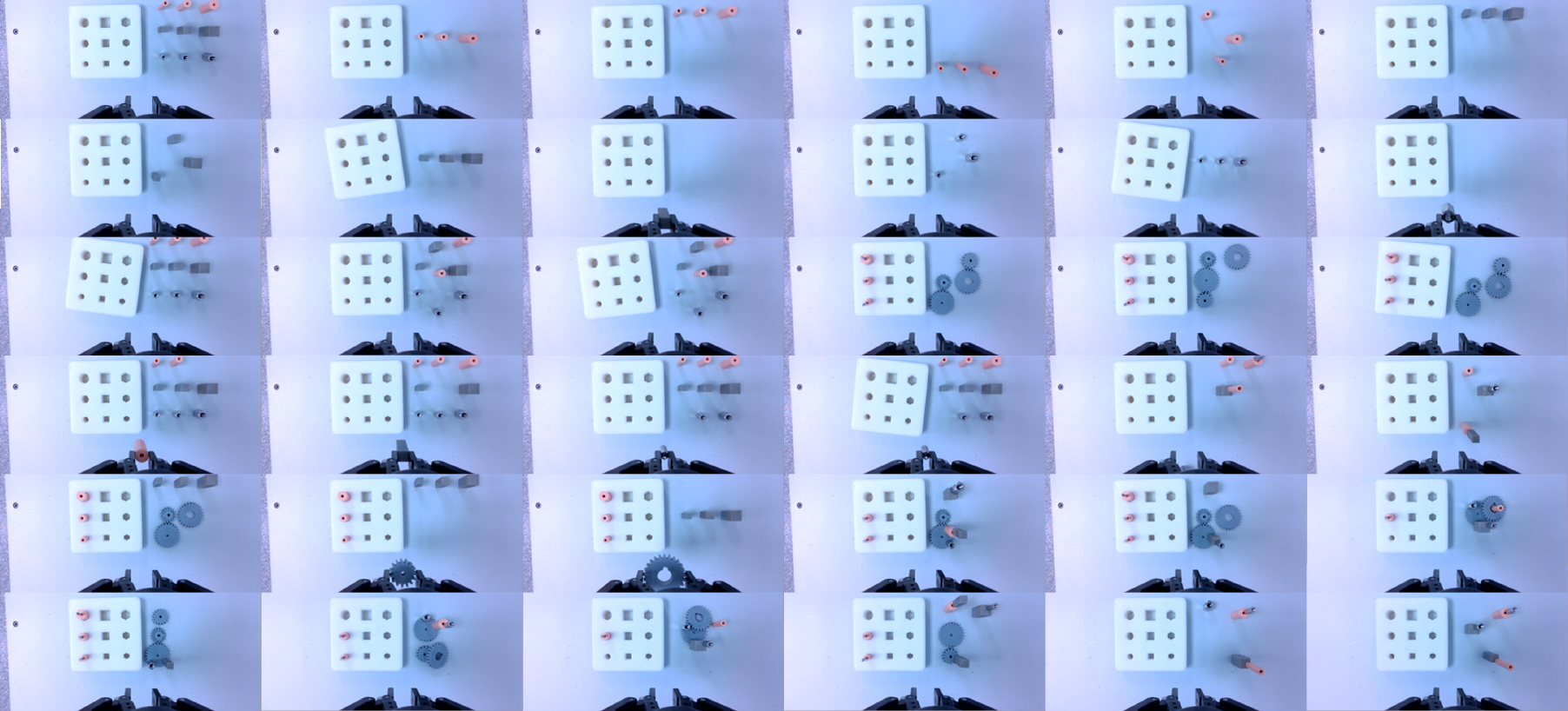}
\caption{Examples of multi-pair assembly scenes.}
\label{fig:real_ass_scenes}
\end{figure}

General evaluation set has 100 sampled multi-pair assembly scenes (77 independent, 23 dependent) with 510 assembly goals, as in Fig.~\ref{fig:real_ass_scenes}. Each goal corresponds to a contact-rich assembly task, specified by a three-role pair $\langle m_j, f_j, \mathrm{tools}_j \rangle$. The scenes cover three peg shapes (circle, hexagon, square) and two gear structures (gear-on-shaft, gear-on-mounted-gear), with varied initial states and stacking dependencies.

The supportive-action discriminator uses 222 training tasks from 96 scenes, 54 validation tasks from 13 scenes, and a single canonical circular peg-in-hole reference. Semantic differences are measured relative to this reference, with supervision assigned according to action applicability. \texttt{align} is supervised on square and hexagonal peg-in-hole tasks to focus learning on semantic variations relevant to angular correction. Circular tasks retain \texttt{release} and \texttt{remove} supervision, while \texttt{align} is masked during training and validation. Training and validation contain no gear tasks. Evaluation covers all 510 tasks (459 peg and 51 gear) before planning guards, with positive/negative counts of 78/432 for \texttt{release}, 50/460 for \texttt{remove}, and 64/446 for \texttt{align}. Unnecessary \texttt{align} predictions on circular and evaluated gear tasks count as false positives. Training and validation records are separate from evaluation records.

\begin{figure}[!t]
\centering
\subfloat[]{
\includegraphics[width=0.40\columnwidth]{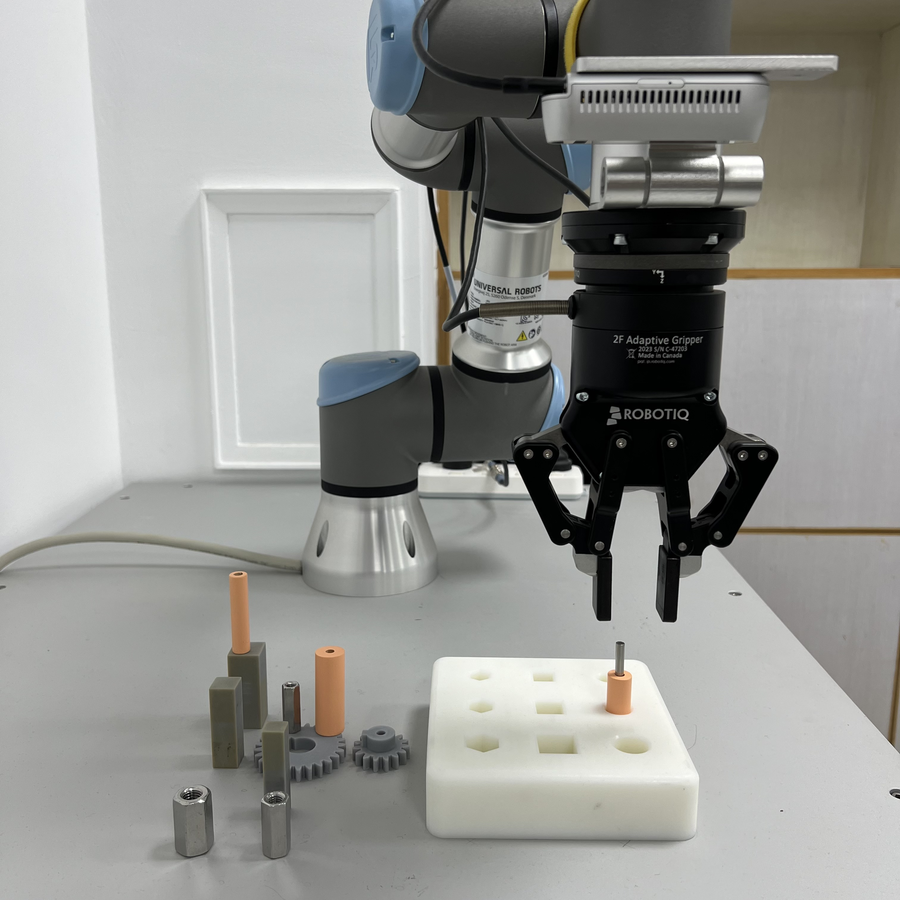}\label{fig:exp_real_p}}\hfill
\subfloat[]{
\includegraphics[width=0.40\columnwidth]{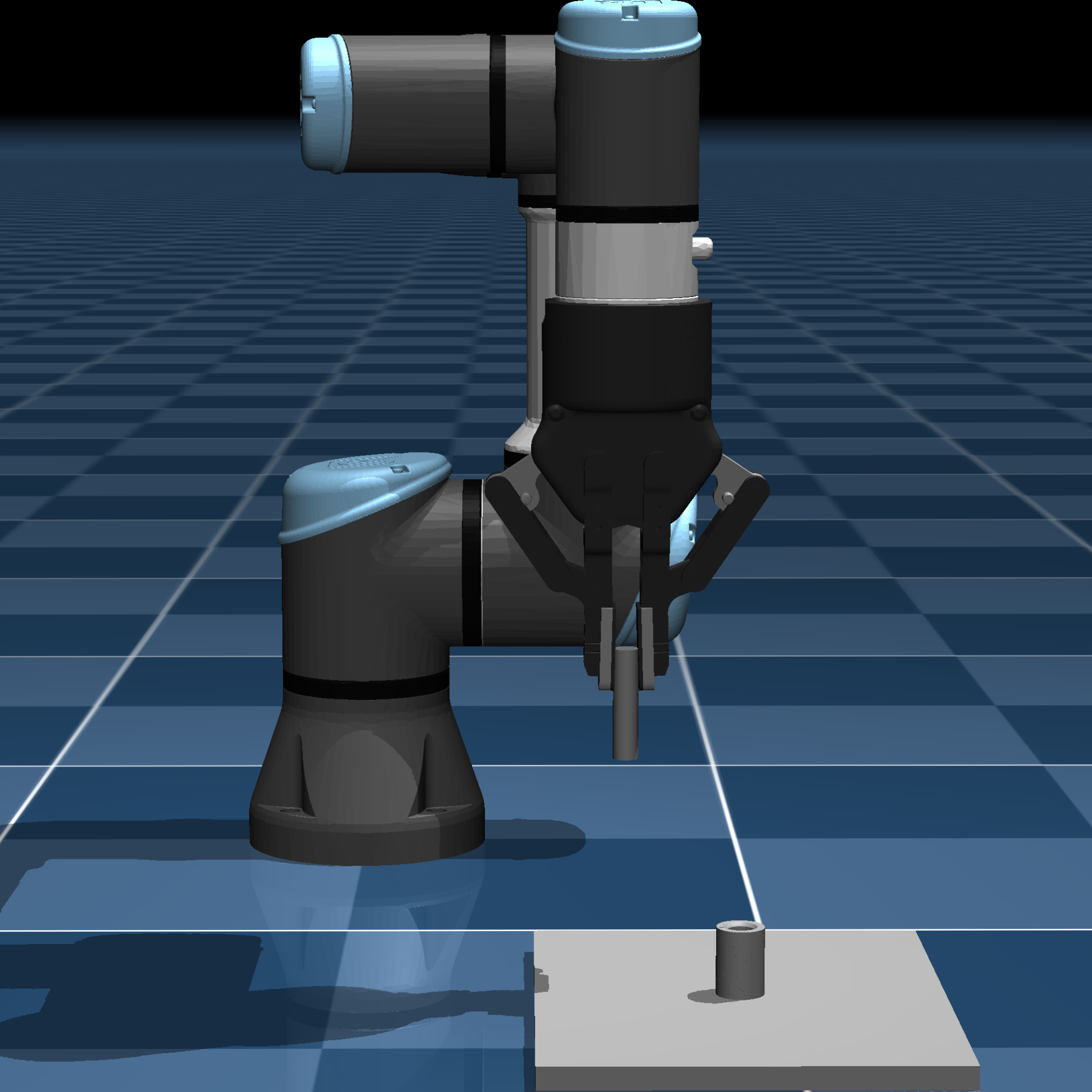}\label{fig:exp_sim_p}}
\caption{Robotic platform.\ (a) Real world execution.\ (b) Physics simulation.}
\label{fig:exp_platforms}
\end{figure}

The robotic platform uses a Universal Robots UR3 CB-series 6-DOF manipulator, Robotiq 2F-85 gripper, Robotiq FT 300-S 6-D F/T sensor, and end-effector-mounted Intel RealSense D435i RGB-D Eye-on-Hand camera at 30~Hz, as shown in Fig.~\ref{fig:exp_real_p}. A standard Linux workstation runs Ubuntu 22.04 and ROS 2 Humble. The hybrid force-control skill combines Archimedean spiral search and PPO impedance control trained in MuJoCo 3.3.2~\cite{todorov_mujoco_2012}, as shown in Fig.~\ref{fig:exp_sim_p}.

\subsection{Experimental Design and Cache-Replay Protocol}
\label{sec:setup_protocol}
The cache-replay protocol reuses saved upstream outputs to keep unchanged stages fixed during controlled comparisons. All planners receive the same Stage-1 perception outputs and assembly goals.

The framework is denoted C0 for comparison with four adapted planners and A0 for ablations and extensions:
\begin{itemize}
\item \textbf{C1. HTN Decomposer}~\cite{nau_shop2_2003}: deterministic basic-action decomposition replaces C0's LLM and sanity/retry loop, retaining the same supportive-action augmentation and topological coordination.
\item \textbf{C2. PDDL Planner}~\cite{fox_pddl21_2003}: classical global state-space planning based on STRIPS/PDDL. Unified Planning 1.3.0/Pyperplan 2.1 uses weighted A* with weight 5 and an additive heuristic.
\item \textbf{C3. LLM-as-BT Adaptation}~\cite{ao_llm-as-bt-planner_2025}: LLM-based task and behavior-tree generation with symbolic simulation feedback. High-level and per-pair BT generation each permit five attempts.
\item \textbf{C4. GRHP-inspired Adaptation}~\cite{li_grhp_2026}: GNN encoders fuse scene and task graphs for a recurrent pointer decoder. Selected semantics and geometry replace the original visual encoders.
\end{itemize}
C4 uses five folds of 64 training, 16 validation, and 20 test scenes, with validation checkpoint selection.

LLM configurations use \texttt{gpt-4.1}, temperature and top-p of 1, a 4096-token output limit, and no fixed LLM seed.

The A0 pipeline comprises P0$\rightarrow$D0$\rightarrow$S0$\rightarrow$T0: (i) \textbf{P}erception frontend, (ii) local task \textbf{D}ecomposition, (iii) \textbf{S}upportive action augmentation, and (iv) global \textbf{T}opological sorting. Table~\ref{tab:ablation_config} specifies the ablation variants and controlled changes.

\begin{table*}[!t]
\caption{Ablation Study Configurations and Algorithm Architecture Reconstructions}
\label{tab:ablation_config}
\centering
\footnotesize
\renewcommand{\arraystretch}{1.25}
\setlength{\tabcolsep}{3pt}
\begin{tabular*}{\textwidth}{@{\extracolsep{\fill}}>{\raggedright\arraybackslash}p{0.055\textwidth}>{\raggedright\arraybackslash}p{0.19\textwidth}>{\raggedright\arraybackslash}p{0.22\textwidth}>{\raggedright\arraybackslash}p{0.46\textwidth}@{}}
\hline
\textbf{Exp. ID} & \textbf{Core Comparison Dimension} & \textbf{Involved Stage / Configuration} & \textbf{Configuration and Controlled Changes} \\
\hline
A0 & Baseline Neuro-Symbolic Architecture & P0$\rightarrow$D0$\rightarrow$S0$\rightarrow$T0 & 11-D semantic differences drive the compact MLP. Guarded augmentation precedes heuristic cross-pair coordination, redundancy pruning, and staged ordering. \\
A1 & Perception degradation & P0, P1 & Fine-tuned CLIP (P0) is replaced by pretrained CLIP (P1) before LLM context construction. \\
A2 & Basic \& Supportive action decomposition & D0$\rightarrow$S0$\rightarrow$T0\newline D1$\rightarrow$S5$\rightarrow$T0 (poor) & Expand the LLM vocabulary to $\mathcal{A}_{\text{base}}\cup\mathcal{A}_{\text{supp}}$ for supportive-action generation and bypass learned augmentation (S5). \\
A3.1 & Augmentation feature representation & D0$\rightarrow$\{S0, S1.1, S1.2\} & Replace local semantics with three-role CLIP embedding differences (S1.1) or prompt-score differences with amplitude summaries (S1.2). MLP dimensions and capacities vary by representation. \\
A3.2 & Augmentation feature distance & D0$\rightarrow$\{S0, S2.1, S2.2\} & Replace only the three JSD summaries by L1 (S2.1) or symmetric KL (S2.2), retaining eight signed probability differences and the MLP. \\
A3.3 & Augmentation discriminator & D0$\rightarrow$\{S0, S3.1, S3.2, S3.3\} & Single-token Transformer (S3.1), wide MLP (S3.2), or logistic regression (S3.3). Same 11-D input, tasks, training and selection. \\
A3.4 & Augmentation extension & D0$\rightarrow$\{S0, S4.1, S4.2\}\newline D2$\rightarrow$\{S4.1, S4.2\} & Frozen S0 for gear transfer and scene variations. \texttt{upright} rules (S4.1) or learned four-action head (S4.2), with frozen (D0) or adapted posture prototypes (D2). \\
A4 & Topological sorting mechanism & D0$\rightarrow$S0$\rightarrow$\{T0, T1, T2\}\newline D1$\rightarrow$S5$\rightarrow$T0 (poor) & T1 disables cross-pair inference. T2 replaces staged sorting with node-level sorting. S5 supplies expanded-action LLM graphs. \\
\hline
\end{tabular*}

\end{table*}

The three-action discriminator comparisons share the same action-specific supervision, validation, and evaluation rules. Decision thresholds are validation-selected and frozen per model and training seed.

Strict complete-scene success requires a valid, nonempty action sequence that completes all assembly goals. Missing outputs or unconfirmed completion count as failures. We report output coverage and confirmed goals separately, with fixed denominators of 100 scenes and 510 goals. Supplied action events are not verified execution lengths. Recovery is the fraction of recorded first failures subsequently accepted within the native retry budget.

Across comparison and ablation experiments, stochastic results are reported as mean $\pm$ sample SD over 5 independent generation runs or training seeds unless otherwise stated (Table~\ref{tab:repetition_structure}). After averaging repetitions within scenes, exploratory complete-scene inference uses 100 paired scenes, pointwise 95\% percentile-bootstrap CIs (10,000 resamples), and two-sided null-centered bootstrap-t tests (99,999 resamples), with Holm correction across five comparisons. C2 uses exact McNemar.

\begin{table}[!t]
\caption{Planning Repetitions and Success-Rate Statistics}
\label{tab:repetition_structure}
\centering
\footnotesize
\renewcommand{\arraystretch}{1.15}
\setlength{\tabcolsep}{3pt}
\begin{tabular*}{\columnwidth}{@{\extracolsep{\fill}}llcc@{}}
\hline
\textbf{Exp. ID} & \textbf{Repetitions} & \textbf{95\% CI (\%)} $\uparrow$ & $\boldsymbol{p}_{\mathrm{Holm}}$ \\
\hline
C0 / A0 & 5 LLM runs $\times$ 5 JS-Aug seeds & \textbf{[90.0, 99.0]} & --- \\
C1 & 5 JS-Aug seeds & \textbf{[90.0, 99.0]} & NA$^\dagger$ \\
C2 & Once & [88.0, 98.0] & 1.00000 \\
C3 & 5 LLM runs & [84.2, 95.2] & 0.14442 \\
C4 & 5 training seeds & [75.0, 88.0] & 0.00464 \\
A2 & 5 LLM runs & [61.4, 73.0] & 0.00005 \\
\hline
\end{tabular*}
\par\smallskip
{\scriptsize\raggedright\textit{Note:} CIs: each method's complete-scene success. $p_{\mathrm{Holm}}$: five comparisons with C0/A0. Runs are averaged within seeds before mean $\pm$ SD across seeds. C4 pools test folds within seeds. $^\dagger$Identical C0/C1 outcomes give a degenerate paired test.\par}
\end{table}

\section{Experimental Results}
\subsection{Comparison of Adapted Planning Methods}

Tables~\ref{tab:repetition_structure} and~\ref{tab:global_exec} compare classical, LLM-based, and graph-learning routes in terms of how they construct assembly plans and complete the evaluated tasks. C0's complete-scene success, confirmed goals, and output coverage are unchanged across all 25 combinations (5 LLM runs $\times$ 5 JS-Aug seeds). After Holm correction, the C0--C4 difference is significant, whereas C0--C2 and C0--C3 are not.

C1 provides a strong reference through hand-authored basic-action decomposition with shared supportive-action augmentation and global coordination. C0 uses the LLM as a language-based bridge from CLIP-derived semantic descriptions to basic-action graphs, without a dedicated HTN decomposition library. Despite the risk of logical errors in LLM-generated plans, C0 matches C1's 95\% complete-scene success with shared downstream stages.

\begin{table}[!t]
\caption{Strict Complete-Scene Success on 100 Recorded Scenes}
\label{tab:global_exec}
\centering
\scriptsize
\renewcommand{\arraystretch}{1.25}
\setlength{\tabcolsep}{1.5pt}
\begin{tabular*}{\columnwidth}{@{\extracolsep{\fill}}lccccc@{}}
\hline
 & \textbf{Conf. } & \textbf{Avg. } & \multicolumn{3}{c}{\textbf{Success (\%) $\uparrow$}} \\[-3.1pt]
\textbf{Method} & \textbf{Goals $\uparrow$} & \textbf{Events $\downarrow$} & \textbf{Overall} & \textbf{Indep.} & \textbf{Dep.} \\
\hline
C0 & \textbf{474} $\pm$ 0.00 & 20.55 $\pm$ 0.00 & \textbf{95.00} $\pm$ 0.00 & \textbf{93.51} $\pm$ 0.00 & \textbf{100.00} $\pm$ 0.00 \\
C1 & \textbf{474} $\pm$ 0.00 & 20.55 $\pm$ 0.00 & \textbf{95.00} $\pm$ 0.00 & \textbf{93.51} $\pm$ 0.00 & \textbf{100.00} $\pm$ 0.00 \\
C2$^{\dagger}$ & 462 & 21.04 & 93.00 (91.00) & 90.91 & \textbf{100.00} (91.30) \\
C3 & 459 $\pm$ 4.30 & \textbf{18.69} $\pm$ 0.18 & 90.00 $\pm$ 1.22 & 89.61 $\pm$ 0.92 & 91.30 $\pm$ 3.07 \\
C4 & 429 $\pm$ 15.51 & 20.49 $\pm$ 0.14 & 81.80 $\pm$ 2.95 & 80.26 $\pm$ 3.83 & 86.96 $\pm$ 0.00 \\
\hline
\end{tabular*}
\par\smallskip
{\scriptsize\raggedright\textit{Note:} Avg. Events: average supplied action events per scene, interpreted alongside success. $^{\dagger}$Parentheses show success rates from the task-order sensitivity analysis.\par}
\end{table}

C2 achieves 93\% overall success through full-scene PDDL planning. C0's augmentation resolves alignment mismatches in two scenes that C2 fails. In two other dependent scenes, C2 succeeds with the favorable original task order but fails under both tested alternatives. When success is required under all three tested orders, C0/C1 succeed in 23/23 dependent scenes and C2 in 21/23, overall success
remains 95\% for C0/C1 and decreases to 91\% for C2.

C3 achieves 90\% complete-scene success with LLM-based ordering and behavior-tree construction. C4 reaches 81.8\% with its GRHP-inspired graph-learning adaptation. These results support the evaluated combination of semantic local construction and explicit constraint handling, with component effects examined separately in A2 and A4 in Section~\ref{sec:core_ablation}. C0 uses a compact, reusable prompt. Across five runs on the 100 scenes, its cumulative input usage is 1.86M tokens, compared with 3.53M for A2 and 9.85M for C3, including retries. Table~\ref{tab:table3} shows that most C0 local graphs pass sanity checks on the first attempt, and retries recover the remaining local failures. C3 recovers most initial failures, but its retry budget does not eliminate scene failures. Full local acceptance won't guarantee complete-scene success. In all five of C0's failed scenes, incorrect rotation semantics from perception enter JS-Aug, which omits required alignment actions.

\subsection{Mechanistic Attribution: Ablation \& Extension Study}
\label{sec:core_ablation}
The ablation study examines perception, the division of basic and supportive reasoning, supportive-action discrimination, and global coordination. We distinguish matched upstream comparisons from changes to the whole pipeline and report local prediction separately from strict scene success. Using the stage-wise cache-replay protocol, controlled ablations and extensions are performed on the baseline architecture A0. 

\begin{table}[!t]
\caption{First Outcomes and Recovery Across Five Generation Runs}
\label{tab:table3}
\centering
\footnotesize
\renewcommand{\arraystretch}{1.15}
\setlength{\tabcolsep}{3pt}
\begin{tabular*}{\columnwidth}{@{\extracolsep{\fill}}lcc@{}}
\hline
\textbf{Metric} & \textbf{C0} & \textbf{C3} \\
Unit per run & 510 pairs & 100 scenes \\
\hline
First success (\%) $\uparrow$ & 95.33 $\pm$ 0.97 & 64.40 $\pm$ 3.51 \\
Final success (\%) $\uparrow$ & 100.00 $\pm$ 0.00 & 90.00 $\pm$ 1.22 \\
Recovery (\%) $\uparrow$ & 100.00 $\pm$ 0.00 & 71.55 $\pm$ 5.14 \\
\hline
First success / all $\uparrow$ & 2431/2550 & 322/500 \\
Final success / all $\uparrow$ & 2550/2550 & 450/500 \\
Recovered / first failures $\uparrow$ & 119/119 & 128/178 \\
\hline
\end{tabular*}
\par\smallskip
{\scriptsize\raggedright\textit{Note:} C0: pair-level sanity acceptance. C3: complete-scene success with attempt budgets of 1 and 5. Recovery uses per-run rates.\par}
\end{table}

\subsubsection{Perception Model Degradation (A1)}

The reliability of high-level symbolic reasoning depends on the quality of low-level semantic tokens. The baseline A0 uses the fine-tuned CLIP model to produce structured mutually exclusive group probabilities $p_{t,i,g}(\ell)$. Fine-tuning employs a low-cost dataset of 133 specifically sampled assembly scenes. In A1, the semantic anchors in the world state $w_{t,j}$ are replaced with pretrained CLIP outputs.

\begin{figure}[!t]
\centering
\includegraphics[width=\columnwidth]{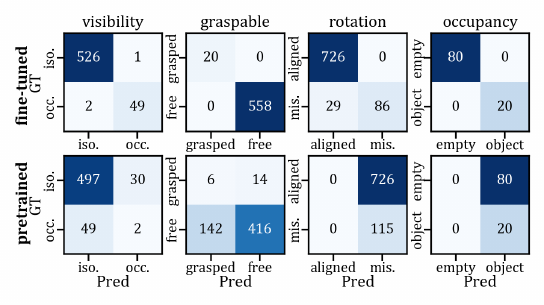}
\caption{Confusion matrices for fine-tuned A0 (top) and pretrained A1 (bottom) across four semantic groups.}
\label{fig:conf_matrices}
\end{figure}

\begin{figure}[!t]
\centering
\includegraphics[width=\columnwidth]{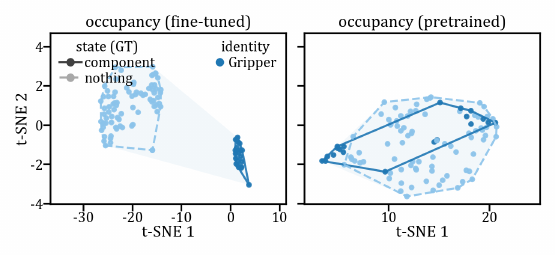}
\caption{Qualitative t-SNE visualization of occupancy for fine-tuned A0 (left) and pretrained A1 (right).}
\label{fig:tsne_vis}
\end{figure}

Table~\ref{tab:ablation_a1_1} reports classification accuracy on the annotated ROIs from the general 100-scene evaluation. Fig.~\ref{fig:conf_matrices} shows corresponding confusion matrices. The pretrained model performs poorly on the challenging fine-grained pose determination (rotation) and gripper occupancy determination (occupancy), predicting a single label throughout each group. Fine-tuning produces large accuracy gains in these two groups and lowers negative log-likelihood (NLL) and Brier scores across all four groups. The t-SNE visualizations in Fig.~\ref{fig:tsne_vis} further demonstrate clearer separation between occupancy states after fine-tuning, whereas the pretrained projection shows substantial overlap.

\begin{table*}[!t]
\caption{Ablation A1: Classification Performance on Mutually Exclusive Semantic Groups Using Fine-Tuned versus Pretrained CLIP}
\label{tab:ablation_a1_1}
\centering
\footnotesize
\renewcommand{\arraystretch}{1.25}
\setlength{\tabcolsep}{5pt}
\begin{tabular}{c c c c c c c c}
\hline
\textbf{Group} & \textbf{$n$} & \textbf{Acc. $\uparrow$} \textbf{(fine-tuned)} & \textbf{Acc. $\uparrow$} \textbf{(pretrained)} & \textbf{$\Delta$Acc. $\uparrow$} & \textbf{$\Delta$NLL $\downarrow$} & \textbf{$\Delta$Brier $\downarrow$} & \textbf{GT-signed margin $\uparrow$} \textbf{(fine-tuned / pretrained)} \\
\hline
visibility & 578 & \textbf{0.995} & 0.863 & +0.131 & -0.375 & -0.240 & \textbf{3.642} / 0.827 \\
graspable & 578 & \textbf{1.000} & 0.730 & +0.270 & -0.581 & -0.402 & \textbf{4.821} / 0.257 \\
rotation & 841 & \textbf{0.966} & 0.137 & +0.829 & -0.967 & -0.814 & \textbf{3.554} / -0.692 \\
occupancy & 100 & \textbf{1.000} & 0.200 & +0.800 & -1.867 & -1.293 & \textbf{4.885} / -1.332 \\
\hline
\end{tabular}
\par\smallskip
{\scriptsize\raggedright\textit{Note:} $n$: evaluated ROI attributes per model. $\Delta$: fine-tuned minus pretrained. NLL, Brier score, and the ground-truth log-odds margin are averages over these observations.\par}
\end{table*}

\subsubsection{Basic \& Supportive Action Decomposition (A2)}

\begin{figure}[!t]
\centering
\subfloat[]{%
\includegraphics[width=\columnwidth]{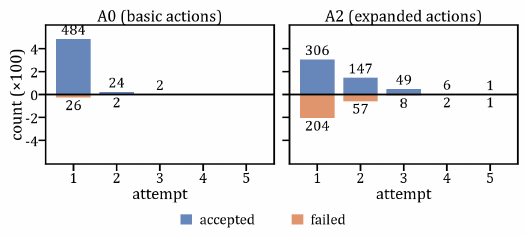}%
}
\par\vspace{0.6em}
\subfloat[]{%
\includegraphics[width=\columnwidth]{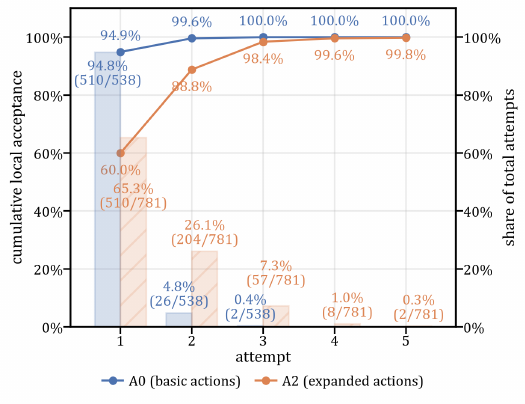}%
}
\caption{Local generation in the first generation run (510 pairs per method). (a) Accepted and failed attempts. (b) Cumulative pair acceptance (left axis) and per-attempt share of all requests (right axis, totaling 538 for A0 and 781 for A2).}
\label{fig:retry_dynamics}
\end{figure}

Basic assembly steps are shared across tasks, whereas supportive actions depend on state. Baseline A0 limits the LLM to basic graph construction and delegates supportive-action reasoning to JS-Aug. A2 expands the action vocabulary from $\mathcal{A}_{\mathrm{base}}$ to $\mathcal{A}_{\mathrm{base}}\cup\mathcal{A}_{\mathrm{supp}}$ and bypasses augmentation, while retaining the same global coordinator. This comparison evaluates the division of reasoning responsibilities rather than the discriminator.

\begin{table}[!t]
\caption{Ablation A2: Local Generation and Complete-Scene Outcomes}
\label{tab:ablation_a2_2}
\centering
\scriptsize
\renewcommand{\arraystretch}{1.15}
\setlength{\tabcolsep}{1.5pt}
\begin{tabular*}{\columnwidth}{@{\extracolsep{\fill}}lcc@{}}
\hline
\textbf{Metric} & \textbf{A0} & \textbf{A2} \\
\hline
\multicolumn{3}{l}{\textit{Local generation}} \\
First-try local acceptance (\%) $\uparrow$ & \textbf{95.33} $\pm$ 0.97 & 58.43 $\pm$ 0.91 \\
Recovered / first failures (\%) $\uparrow$ & \textbf{100.00} $\pm$ 0.00 & 99.34 $\pm$ 0.25 \\
Extra attempts / pair $\downarrow$ & \textbf{0.05} $\pm$ 0.01 & 0.56 $\pm$ 0.02 \\
Above annotated length (\%) & 0.00 $\pm$ 0.00 & 6.49 $\pm$ 1.88 \\
Mean annotated length gap & 0.00 $\pm$ 0.00 & 0.02 $\pm$ 0.02 \\
Redundant support (template) (\%) $\downarrow$ & \textbf{0.00} $\pm$ 0.00 & 6.13 $\pm$ 1.61 \\
Final local acceptance (\%) $\uparrow$ & \textbf{100.00} $\pm$ 0.00 & 99.73 $\pm$ 0.11 \\
Recovered / all pairs (\%) & 4.67 $\pm$ 0.97 & 41.29 $\pm$ 0.86 \\
Raw-to-template length reduction (\%) & 2.08 $\pm$ 0.17 & 4.39 $\pm$ 0.20 \\
\hline
\multicolumn{3}{l}{\textit{Complete-scene evaluation}} \\
Scene success (\%) $\uparrow$ & \textbf{95.00} $\pm$ 0.00 & 67.20 $\pm$ 5.22 \\
Available outputs / 100 $\uparrow$ & \textbf{100.00} $\pm$ 0.00 & 98.60 $\pm$ 0.55 \\
Confirmed goals / 510 $\uparrow$ & \textbf{474.00} $\pm$ 0.00 & 398.80 $\pm$ 22.65 \\
\hline
\end{tabular*}
\par\smallskip
{\scriptsize\raggedright\textit{Note:} Per-run averages use $\max(0,1-L_{\mathrm{template}}/L_{\mathrm{raw}})$ over available pairs for reduction, and known redundant/total support actions over 510 pairs for redundancy (zero with no support). Gap: template length minus annotated length. Missing templates (7/2,550): excluded from length metrics, zero redundancy, counted as failures.\par}
\end{table}

\begin{figure*}[!t]
\centering
\includegraphics[width=\textwidth]{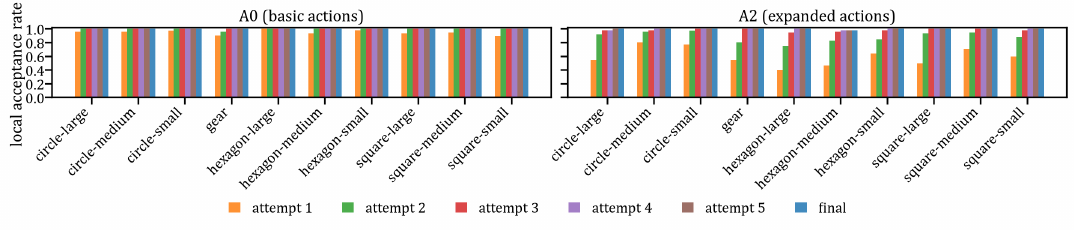}
\caption{Cumulative local acceptance by group in the first generation run, for A0 (left) and A2 (right). Nine peg shape-size groups contain 459 pairs. Gear pools 51 pairs (10 gear--gear and 41 gear--shaft) without size binning.}
\label{fig:shape_size_success}
\end{figure*}

As shown in Table~\ref{tab:ablation_a2_2}, expanding the action vocabulary in A2 lowers first-try local acceptance from 95.33\% to 58.43\% and increases the average extra attempts per pair from 0.05 to 0.56. Feedback recovers most initial failures, bringing final local acceptance to 100\% for A0 and 99.73\% for A2. However, complete-scene success remains 95\% and 67.20\%, respectively, with a significant paired difference after Holm correction in Table~\ref{tab:repetition_structure}.  Thus, successful local repair does not ensure complete-scene execution.

\begin{figure}[!t]
\centering
\includegraphics[width=\columnwidth]{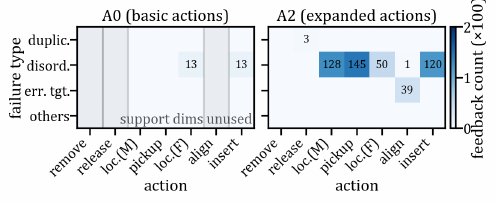}
\caption{Validation feedback by error type and action in the first generation run. A0/A2 have 28/272 failed attempts and 26/486 feedback items. An attempt may contribute multiple items. Gray columns mark unused supportive-action dimensions in A0.}
\label{fig:failure_heatmap}
\end{figure}

\begin{figure}[!t]
\centering
\includegraphics[width=\columnwidth]{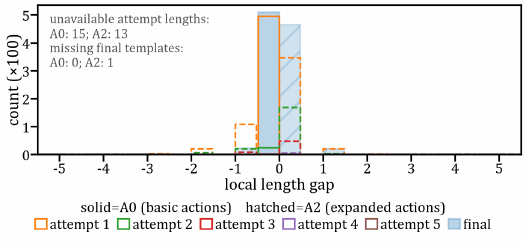}
\caption{Local length gaps in the first generation run. Final templates are filled (A0 solid, A2 hatched), while attempt histograms use solid/dashed outlines. Unavailable lengths and missing final templates (gray counts) are excluded, while failures remain in acceptance and retry statistics.}
\label{fig:pruning_evolution}
\end{figure}

In the first generation run, Figs.~\ref{fig:retry_dynamics} and~\ref{fig:shape_size_success} show that A2 relies more heavily on retries across groups. Its validation feedback concentrates on action-order and target errors in Fig.~\ref{fig:failure_heatmap}. Raw and template lengths count actions before and after sanity checking. The annotated length counts basic actions before JS-Aug for A0 and additionally the necessary \texttt{release}, \texttt{remove}, and \texttt{align} actions for A2. Fig.~\ref{fig:pruning_evolution} shows both positive and negative deviations for A2. On average across five generation runs, 6.49\% of its available templates exceed the annotated length and 4.80\% fall below it. The near-zero mean gap partly reflects cancellation. Matching the annotated length establishes neither final executability nor optimality.

These results indicate that purely LLM-driven supportive action generation suffers from structural inefficiencies without the separate JS-Aug augmentation stage. The shape-size results further suggest that A0 generalizes better than A2 across the evaluated variations in geometric appearance. Limiting the LLM to generic logic of basic graph construction and delegating state-dependent supportive reasoning to JS-Aug improves first-try acceptance, reduces retries, and yields higher complete-scene success in the evaluated tasks. These results support the proposed division of reasoning responsibilities.

\subsubsection{Augmentation Design and Extension (A3)}

A3 first examines how representation, distance summaries, discriminator capacity, and supervision affect supportive-action prediction. Table~\ref{tab:ablation_a3} summarizes these comparisons on the general 100-scene evaluation set using five discriminator training seeds (101, 202, 303, 404, 505). Balanced accuracy (BA) averages recall for the positive and negative classes, while false-positive rate (FPR) measures unnecessary activation among negative tasks. We then examine reuse across assembly structures or scene layouts, and extension to a new supportive action.

\textit{Feature representation (A3.1):} We compare S0's local semantic representation with three-role CLIP embedding differences (S1.1) and prompt-score differences (S1.2). S0 combines three JSD summaries with eight signed probability differences relative to the reference. Its higher micro- and macro-F1, together with improved \texttt{remove} and \texttt{align} BA over S1.2, support the task-related semantic representation.

\begin{table*}[!t]
\caption{Ablation A3: Discriminative Power of Feature, Distance, and Discriminator Variants}
\label{tab:ablation_a3}
\centering
\scriptsize
\renewcommand{\arraystretch}{1.25}
\setlength{\tabcolsep}{1.9pt}
\begin{tabular*}{\textwidth}{@{\extracolsep{\fill}}lccccccccccc@{}}
\hline
\textbf{Config} & \textbf{Dist.} & \textbf{Disc.} & \textbf{Param.} & \textbf{\boldmath micro-F1 (\%) $\uparrow$} & \textbf{\boldmath macro-F1 (\%) $\uparrow$} & \multicolumn{3}{c}{\textbf{\boldmath 45-task BA (\%) $\uparrow$}} & \multicolumn{3}{c}{\textbf{\boldmath 222-task BA (\%) $\uparrow$}} \\
 &  &  &  &  &  & \textbf{\texttt{release}} & \textbf{\texttt{remove}} & \textbf{\texttt{align}} & \textbf{\texttt{release}} & \textbf{\texttt{remove}} & \textbf{\texttt{align}} \\
\hline
S0 & JSD & MLP & 2439 & \textbf{93.48} $\pm$ 0.15 & \textbf{92.59} $\pm$ 0.18 & \textbf{100.00} $\pm$ 0.00 & \textbf{98.74} $\pm$ 0.46 & \textbf{85.52} $\pm$ 0.70 & \textbf{100.00} $\pm$ 0.00 & 99.22 $\pm$ 0.05 & 86.56 $\pm$ 0.35 \\
S1.1 & --- & MLP$^{\dagger}$ & 12069255 & 69.20 $\pm$ 4.33 & 72.78 $\pm$ 2.55 & --- & --- & --- & \textbf{100.00} $\pm$ 0.00 & 93.23 $\pm$ 1.45 & 77.36 $\pm$ 4.27 \\
S1.2 & $\overline{|\Delta\mathrm{score}|}$ & MLP & 2439 & 71.92 $\pm$ 12.74 & 80.92 $\pm$ 7.11 & 97.39 $\pm$ 2.18 & 87.32 $\pm$ 8.19 & 70.77 $\pm$ 1.48 & \textbf{100.00} $\pm$ 0.00 & 97.78 $\pm$ 0.00 & 75.22 $\pm$ 6.73 \\
S2.1 & L1 & MLP & 2439 & 92.64 $\pm$ 0.39 & 91.60 $\pm$ 0.45 & --- & --- & --- & \textbf{100.00} $\pm$ 0.00 & 99.00 $\pm$ 0.14 & 85.62 $\pm$ 0.43 \\
S2.2 & Sym. KL & MLP & 2439 & 92.77 $\pm$ 0.57 & 91.69 $\pm$ 0.68 & --- & --- & --- & \textbf{100.00} $\pm$ 0.00 & \textbf{99.28} $\pm$ 0.18 & 84.22 $\pm$ 0.65 \\
S3.1 & JSD & TF & 2467 & 78.02 $\pm$ 6.79 & 80.79 $\pm$ 3.71 & \textbf{100.00} $\pm$ 0.00 & 97.70 $\pm$ 1.38 & 75.92 $\pm$ 1.93 & \textbf{100.00} $\pm$ 0.00 & 98.59 $\pm$ 0.20 & 79.07 $\pm$ 4.22 \\
S3.2 & JSD & MLP-W & 9939 & 93.23 $\pm$ 0.28 & 92.31 $\pm$ 0.32 & \textbf{100.00} $\pm$ 0.00 & 98.70 $\pm$ 0.58 & 84.84 $\pm$ 0.91 & \textbf{100.00} $\pm$ 0.00 & 99.09 $\pm$ 0.12 & \textbf{86.72} $\pm$ 0.00 \\
S3.3 & JSD & LR & \textbf{36} & 92.14 $\pm$ 0.49 & 90.86 $\pm$ 0.62 & 99.07 $\pm$ 1.01 & 96.12 $\pm$ 4.73 & 76.08 $\pm$ 4.25 & \textbf{100.00} $\pm$ 0.00 & 98.54 $\pm$ 0.18 & 81.72 $\pm$ 1.05 \\
\hline
\end{tabular*}
\par\vspace{3pt}
{\scriptsize\raggedright\textit{Note:} Param.: discriminator weights and biases. F1 uses the 222-task training budget. 45-task results: averages over three subsets per seed. Overline: average absolute score difference per role. ---: no summary or unevaluated combination. The compact MLP is approximately parameter-matched to TF. $^{\dagger}$Larger MLP for 1536-D embedding differences.\par}
\end{table*}

To assess the contribution of learned mapping, a CLIP-only baseline bypasses the discriminator and uses the visual-prior demand probability for any supportive action $a \in \mathcal{A}_{\mathrm{supp}}$:
\begin{equation}
\label{eq:clip_action_prob}
p_a^{\text{clip}} = \sum_{\ell \in \mathcal{L}_a^+} p_{t,i,g}(\ell),
\end{equation}
where $\mathcal{L}_a^+$ contains the positive semantic labels for $a$. The threshold is 0.5. The semantic margin $m_a=2p_a^{\text{clip}}-1$ defines the high-confusion region $|m_a|<0.4$, equivalent to confusion degree $\chi_a=1-|m_a|>0.6$.

\begin{figure}[!t]
\centering
\includegraphics[width=\columnwidth]{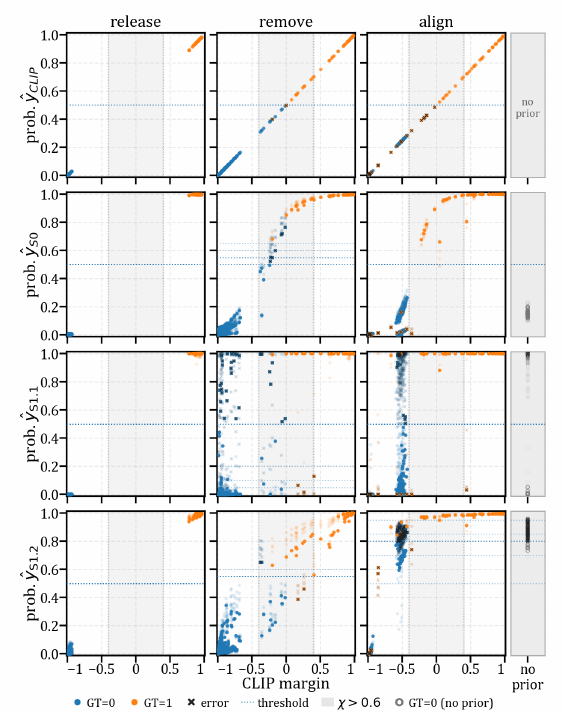}
\caption{Predictive performance versus CLIP margin. Columns correspond to \texttt{release}, \texttt{remove}, and \texttt{align}, and rows compare CLIP-only with the three feature variants. The side strip shows raw outputs for 51 gear tasks lacking CLIP \texttt{align} priors, scored as known negatives in the 510-task discriminator evaluation. Seed 101 is highlighted, with four other seeds translucent. Blue lines mark thresholds, crosses mark errors, central shading indicates $\chi>0.6$.}
\label{fig:jstrans_output_prob}
\end{figure}

Fig.~\ref{fig:jstrans_output_prob} shows distinct probability patterns across the CLIP margin. S0 exhibits a more regular distribution, with most positive tasks concentrated at high probabilities and negative tasks at low probabilities. S1.1 produces high-confidence false-positive outliers, whereas S1.2 shows a more dispersed distribution. All representations classify \texttt{release} correctly. For \texttt{remove}, S0 recovers CLIP's missed boundary positives and preserves correct decisions outside the high-confusion region, with additional false positives confined to that region. S1.1 also produces high-confidence false positives outside it, whereas S1.2 trades fewer false positives for missed positives. For \texttt{align}, S0 improves detection of boundary positives without activating negative tasks, including circular tasks without direct \texttt{align}-negative supervision. S1.1 and S1.2 detect more positives on average but also produce high-confidence false positives. False negatives persist outside the high-confusion region for CLIP and all three variants, consistent with the visual ambiguity of small rotation angles: a large absolute CLIP margin does not necessarily indicate a correct semantic prediction. Across the five training seeds, S0's probability patterns for \texttt{remove} and \texttt{align} nearly overlap. In contrast, S1.1 and S1.2 exhibit more pronounced vertical variation in output probability for the same tasks.

\textit{Feature distance (A3.2):} For 11-D inputs, we replace the three JSD summaries with L1 (S2.1) or symmetric KL (S2.2), retaining the eight signed probability differences and the same MLP architecture. Table~\ref{tab:ablation_a3} shows higher mean micro- and macro-F1 and \texttt{align} BA with JSD, together with smaller F1 variation across seeds. Symmetric KL nevertheless achieves slightly higher \texttt{remove} BA. JSD thus offers a better trade-off for thresholded action prediction.

\textit{Discriminator (A3.3):} With identical 11-D inputs, tasks, and training/selection policy, S0's compact MLP is compared with a single-token Transformer, wider MLP, and logistic regression (S3.1--S3.3). With 45 supervised local tasks, S0 already achieves \texttt{remove} BA of $98.74\pm0.46\%$ and \texttt{align} BA of $85.52\pm0.70\%$, the highest means among the four discriminator heads for both actions. At 222 tasks, S0 retains the highest mean micro- and macro-F1, supporting the compact MLP as an effective choice for supportive-action prediction.

\begin{figure}[!t]
\centering
\includegraphics[width=88mm]{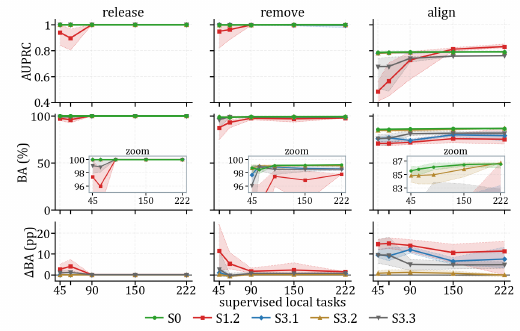}
\caption{Supportive-action learning curves. Rows show trapezoidal AUPRC, BA, and paired $\Delta$BA (S0 minus comparison). Curves and bands show means and ranges across five training seeds, with results averaged over training subsets within each seed.}
\label{fig:linear_probe}
\end{figure}

Across the budgets in Table~\ref{tab:supervision}, Fig.~\ref{fig:linear_probe} shows modest changes in S0's AUPRC and BA with narrow seed ranges. The mean \texttt{align} $\Delta$BA favors S0 over the plotted alternatives, except for S3.2's slight lead at 222 tasks. At that budget, S0 achieves higher \texttt{align} BA than S1.2 despite lower AUPRC, indicating better action prediction at the evaluation threshold.

\begin{table}[!t]
\caption{Supervision Budgets and Task Composition for Ablation A3}
\label{tab:supervision}
\centering
\footnotesize
\renewcommand{\arraystretch}{1.25}
\setlength{\tabcolsep}{2pt}
\begin{tabular*}{\columnwidth}{@{\extracolsep{\fill}}cccccc@{}}
\hline
\textbf{Tasks} & \textbf{Subsets} & \textbf{Scenes} & \textbf{Update per} & \multicolumn{2}{c}{\textbf{Positive / Negative / NA labels}} \\
 & & & \textbf{Epoch} & \textbf{\texttt{release} and \texttt{remove}} & \textbf{\texttt{align}} \\
\hline
45 & 3 & 41$^{\dagger}$ & 1 & 18 / 27 / 0$^{\dagger}$ & 12 / 18 / 15 \\
60 & 3 & 50$^{\dagger}$ & 1 & 24 / 36 / 0 & 17 / 23 / 20$^{\dagger}$ \\
90 & 3 & 64$^{\dagger}$ & 2 & 36 / 54 / 0$^{\dagger}$ & 25 / 35 / 30$^{\dagger}$ \\
150 & 3 & 86$^{\dagger}$ & 3 & 60 / 90 / 0 & 40 / 60 / 50$^{\dagger}$ \\
222 & 1 & 96 & 4 & 89 / 133 / 0 & 60 / 88 / 74 \\
\hline
\end{tabular*}
\par\smallskip
{\footnotesize\raggedright\textit{Note:} Subsets: sampling seeds 7201, 7202, 7203, with the full set at 222 tasks. NA labels are excluded from the training loss. Each budget has equal task counts for the three peg types. Batch size is 64. $^{\dagger}$For varying counts, the most frequent value or positive/negative/NA combination across subsets is shown. A row may combine values from different subsets.\par}
\end{table}

\begin{table}[!t]
\caption{Ablation A3.4: Supportive Action Transfer to Gear Assembly}
\label{tab:gear}
\centering
\scriptsize
\renewcommand{\arraystretch}{1.25}
\setlength{\tabcolsep}{2pt}
\begin{tabular*}{\columnwidth}{@{\extracolsep{\fill}}lcccc@{}}
\hline
\textbf{Gear Role} & \textbf{\texttt{release}} & \textbf{\texttt{remove}} & \textbf{\texttt{align}} & \textbf{\texttt{remove}} \\
 & \textbf{\boldmath BA (\%) $\uparrow$} & \textbf{\boldmath BA (\%) $\uparrow$} & \textbf{\boldmath FPR (\%) $\downarrow$} & \textbf{\boldmath Log-loss $\downarrow$} \\
\hline
Target & 100.00 $\pm$ 0.00 & 99.82 $\pm$ 0.41 & 0.00 $\pm$ 0.00 & 0.0768 $\pm$ 0.0020 \\
Accompany & 99.12 $\pm$ 0.00 & 97.10 $\pm$ 0.00 & 0.00 $\pm$ 0.00 & 0.1208 $\pm$ 0.0150 \\
\hline
\end{tabular*}
\par\vspace{3pt}
{\scriptsize\raggedright\textit{Note:} Target: $n=85$ gear tasks (from general/new set: 51/34). Accompany: $n=154$ peg tasks (52/102). Metrics are computed on combination within each training seed.\par}
\end{table}

\begin{table*}[!t]
\caption{Ablation A3.4: \texttt{upright} Action Extension and Original-Action Performance (Separate Evaluation)}
\label{tab:upright-separate}
\centering
\footnotesize
\renewcommand{\arraystretch}{1.25}
\setlength{\tabcolsep}{5pt}
\begin{tabular*}{\textwidth}{@{\extracolsep{\fill}}lccccc lcc@{}}
\hline
\textbf{Config} & \multicolumn{2}{c}{\textbf{Added Training Data}} & \multicolumn{3}{c}{\textbf{\boldmath Extension Action: \texttt{upright} ($n=218$)}} & \multicolumn{3}{c}{\textbf{\boldmath Original Actions ($n=510$)}} \\
 & \textbf{\boldmath $\Delta$Pairs} & \textbf{\boldmath $\Delta$ROIs} & \textbf{\boldmath BA (\%) $\uparrow$} & \textbf{\boldmath FPR (\%) $\downarrow$} & \textbf{\boldmath Log-loss $\downarrow$} &  & \textbf{\boldmath $\Delta$BA (pp) $\uparrow$} & \textbf{\boldmath $\Delta$FPR (pp) $\downarrow$} \\
\hline
D0$\rightarrow$S4.1$^{\dagger}$ & +0 & +0 & 58.38 & 28.96 & 0.6701 & --- & --- & --- \\
 &  &  &  &  &  & \texttt{release} & \textbf{+0.00} $\pm$ 0.00 & \textbf{+0.00} $\pm$ 0.00 \\
D0$\rightarrow$S4.2 & +66 & +0 & 79.82 $\pm$ 6.74 & 14.64 $\pm$ 7.10 & 0.2877 $\pm$ 0.0196 & \texttt{remove} & \textbf{+0.46} $\pm$ 0.12 & \textbf{-0.91} $\pm$ 0.24 \\
 &  &  &  &  &  & \texttt{align} & \textbf{+0.16} $\pm$ 0.35 & \textbf{+0.00} $\pm$ 0.00 \\
D2$\rightarrow$S4.1$^{\dagger}$ & +0 & +18 & 90.65 & 15.85 & 0.1826 & --- & --- & --- \\
 &  &  &  &  &  & \texttt{release} & \textbf{+0.00} $\pm$ 0.00 & \textbf{+0.00} $\pm$ 0.00 \\
D2$\rightarrow$S4.2 & +66 & +18 & \textbf{95.25} $\pm$ 0.32 & \textbf{2.08} $\pm$ 1.18 & \textbf{0.0855} $\pm$ 0.0025 & \texttt{remove} & -0.35 $\pm$ 0.14 & +0.70 $\pm$ 0.28 \\
 &  &  &  &  &  & \texttt{align} & \textbf{+0.16} $\pm$ 0.35 & \textbf{+0.00} $\pm$ 0.00 \\
\hline
\end{tabular*}
\par\vspace{3pt}
{\scriptsize\raggedright\textit{Note:} $\Delta$BA/$\Delta$FPR are changes from S0 in percentage points. \texttt{upright} labels: 35 positive, 183 negative. --- indicates no original-action output. $^{\dagger}$Rules are evaluated once.\par}
\end{table*}

\begin{figure}[!t]
\centering
\subfloat[]{\includegraphics[width=0.3\columnwidth]{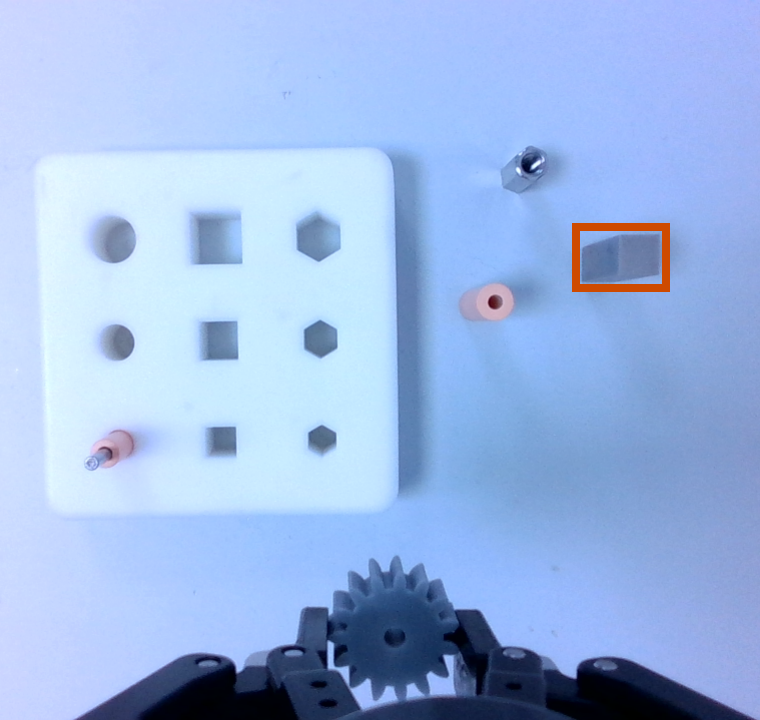}}\hfill
\subfloat[]{\includegraphics[width=0.3\columnwidth]{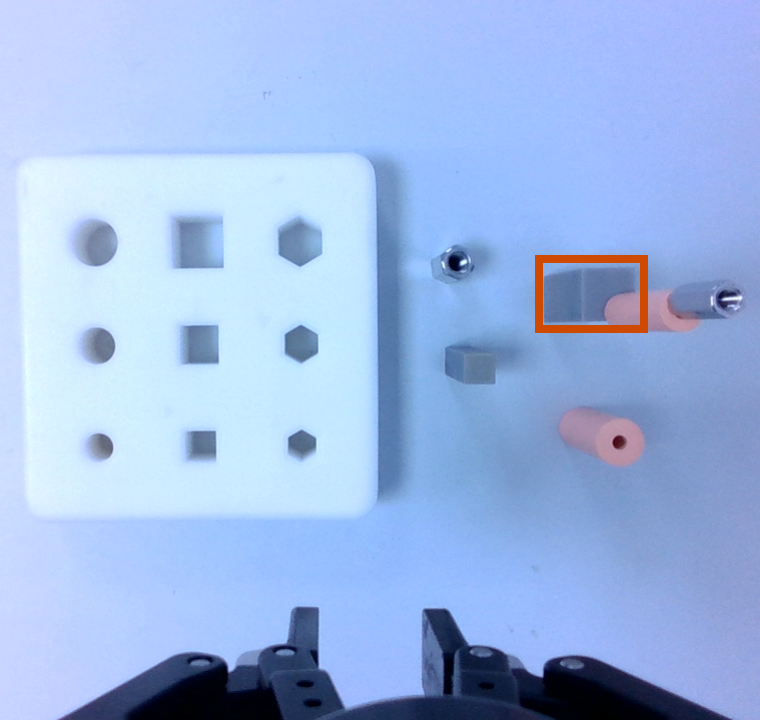}}\hfill
\subfloat[]{\includegraphics[width=0.3\columnwidth]{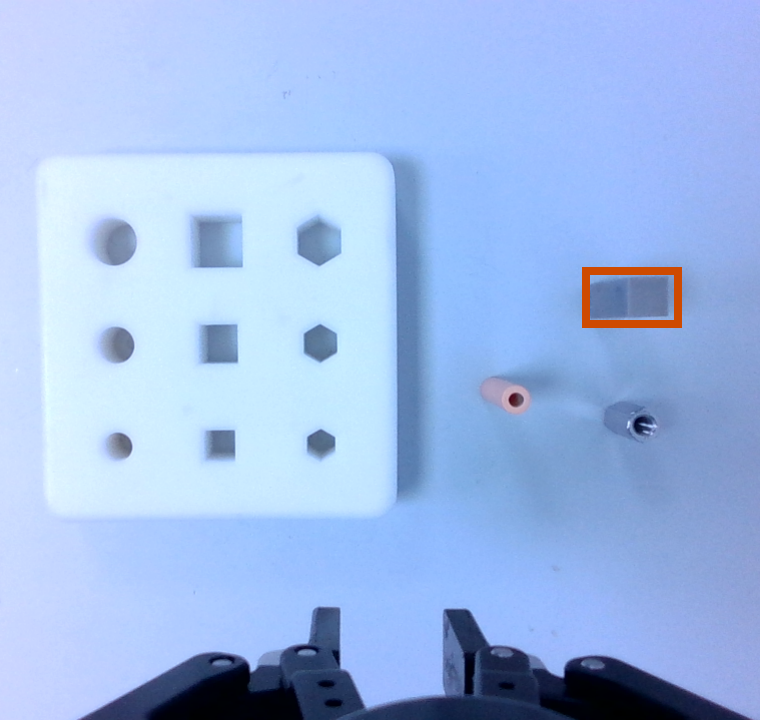}}
\\
\subfloat[]{\includegraphics[width=0.3\columnwidth]{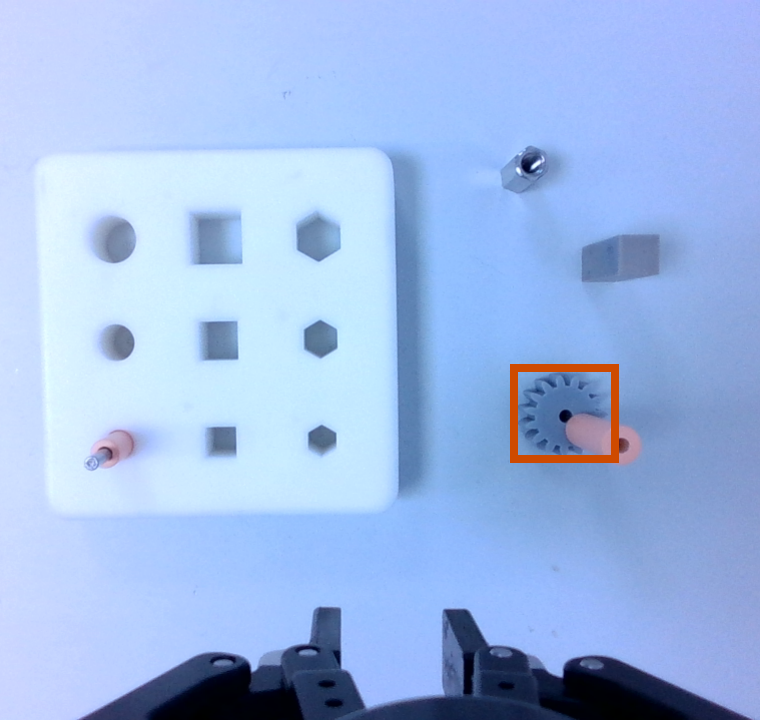}}\hfill
\subfloat[]{\includegraphics[width=0.3\columnwidth]{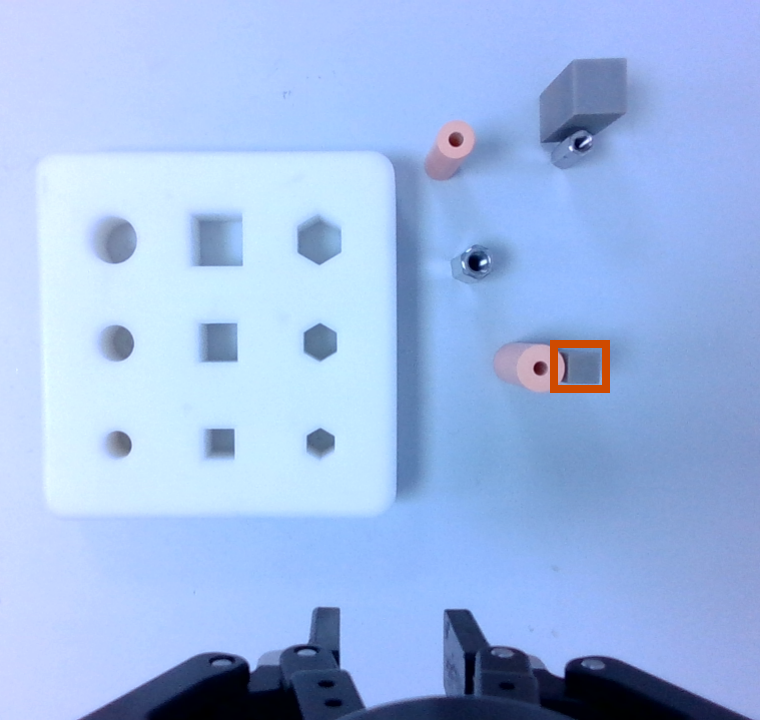}}\hfill
\subfloat[]{\includegraphics[width=0.3\columnwidth]{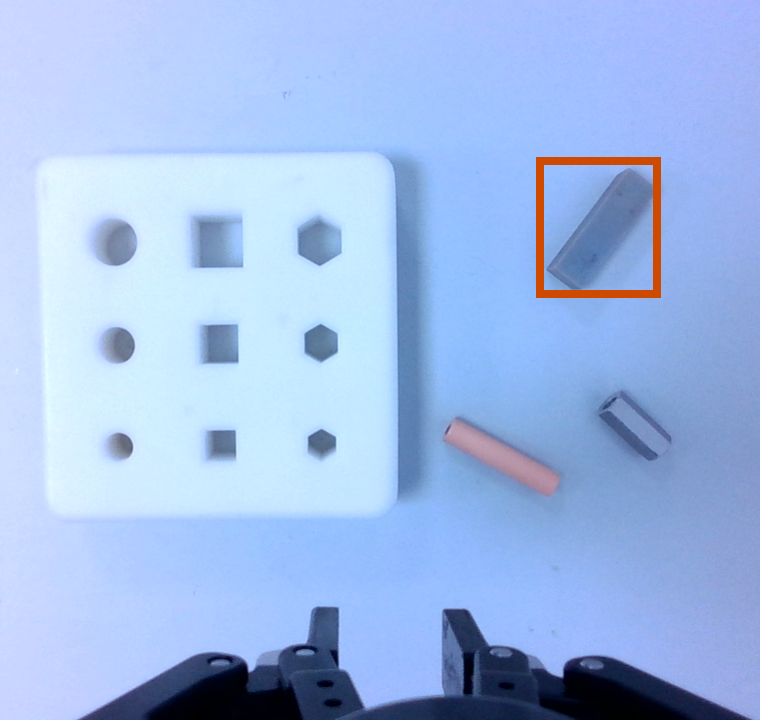}}
\caption{Examples of supportive-action reuse and extension. Left: (a) a peg task with gear occupancy and (d) a target gear task. Middle: (b) overlapping and (e) adjacent layouts sharing a \texttt{remove} requirement. Right: (c) standing and (f) lying postures illustrating the extended \texttt{upright} requirement. Boxes mark example task objects.}
\label{fig:augmentation_extension}
\end{figure}

\textit{Extension (A3.4):} We first test whether existing supportive actions can be reused across assembly structures and changed scene conditions. Both studies use the frozen S0 trained on peg-in-hole tasks in Section~\ref{sec:setup_data}, with their evaluation tasks excluded from training and validation. We then introduce posture as an additional semantic group to extend the action vocabulary with \texttt{upright}. Fig.~\ref{fig:augmentation_extension} illustrates these three settings.

Table~\ref{tab:gear} reports two gear-related settings: gear as the target or accompanying peg tasks. For target gear, female recipient (peg with shaft or mounted gear, both can serve as either male or female) lacks a rotation semantic group, which is filled with the canonical reference probabilities, giving zero rotation-difference input features without forcing \texttt{align} predictions to zero, for all 11-D variants in Table~\ref{tab:ablation_a3} and Fig.~\ref{fig:jstrans_output_prob}. High \texttt{release} and \texttt{remove} BA support reuse of the learned supportive-action mapping across these structures. With all \texttt{align} labels negative, zero FPR indicates no unnecessary activations.

The overlapping and adjacent layouts in Fig.~\ref{fig:augmentation_extension}(b) and (e) differ visually but share a need for \texttt{remove} before grasping. In the adjacent arrangement, a larger neighboring part restricts gripper access to the smaller target despite leaving it visible. S0 correctly predicts both conditions in 4 of 7 layout pairs requiring a change in \texttt{remove} and 13 of 14 requiring none. These results support reuse of learned supportive-action knowledge across some changed layouts through abstract local semantics, without retraining for each specific layout.

The \texttt{upright} extension in Table~\ref{tab:upright-separate} is evaluated on 218 new posture tasks, while the influence on original actions is evaluated on 510 tasks from the general set. S4.1 uses a deterministic posture-threshold rule. S4.2 is retrained on 288 tasks to predict all four actions, with 69 validation tasks for model and threshold selection. D2 adapts two posture prototypes (standing and lying) using 18 fitting ROIs and nine selection ROIs with seed 101, keeping both CLIP encoders frozen. With D0's unadapted posture prototypes, learned augmentation outperforms the rule. Posture adaptation alone raises rule BA to 90.65\%, and learning with adapted inputs further raises BA to $95.25\pm0.32\%$ while reducing FPR from 15.85\% to $2.08\pm1.18\%$. These complementary gains accompany small original-action changes, including a $0.35\pm0.14$-point reduction in \texttt{remove} BA. This demonstrates a supervised action extension, without establishing complete-scene or robot-execution gains.

\subsubsection{Global TOPO Mechanism Restructuring (A4)}

T0, T1, and T2 share S0 inputs. T1 disables cross-pair dependency inference. T2 replaces staged sorting with node-level topological sorting. The separate S5$\rightarrow$T0 condition uses A2's expanded-action graphs and is labeled T0 (poor).

Table~\ref{tab:ablation_a4} shows that T0 achieves 95\% strict scene success, compared with 61\% for T1 and 57\% for T2. T2 retains full output coverage, but adds 38 failures first detected at \texttt{pick} or \texttt{localize} with unmet empty-gripper preconditions. With A2's expanded-action graphs, T0 (poor) achieves an average success rate of 67.20\%. However, across five generation runs, it prunes 324 redundant \texttt{release} and 291 redundant \texttt{remove} steps from local plans, shortening sequences in 158 of 500 scene evaluations, 110 of which achieve complete-scene success. Thus, TOPO can eliminate cross-task redundancy, but cannot compensate for all local errors. T0 also reduces the supplied action sequences, yielding a reduction contribution of 4.59\% compared with 0\% for T1.

\begin{table}[!t]
\caption{Ablation A4: Contribution of Topological Coordination Operators to Global Executability}
\label{tab:ablation_a4}
\centering
\footnotesize
\renewcommand{\arraystretch}{1.25}
\setlength{\tabcolsep}{0.95pt}
\begin{tabular*}{\columnwidth}{@{\extracolsep{\fill}}ccccc@{}}
\hline
\textbf{Config} & \shortstack{\textbf{Scene}$^{\dagger}$\\\textbf{\boldmath Success (\%)$\uparrow$}} & \shortstack{\textbf{Available}$^{\dagger}$\\\textbf{\boldmath Outputs (\%)$\uparrow$}} & \shortstack{\textbf{Confirmed}$^{\ddagger}$\\\textbf{\boldmath Goals (\%)$\uparrow$}} & \shortstack{\textbf{Reduction}$^{\dagger}$\\\textbf{Contribution (\%)}} \\
\hline
S0$\rightarrow$T0 & \textbf{95.00} $\pm$ 0.00 & \textbf{100.00} $\pm$ 0.00 & \textbf{92.94} $\pm$ 0.00 & \textbf{4.59} $\pm$ 0.00 \\
S0$\rightarrow$T1 & 61.00 $\pm$ 0.00 & \textbf{100.00} $\pm$ 0.00 & 67.25 $\pm$ 0.00 & 0.00 $\pm$ 0.00 \\
S0$\rightarrow$T2 & 57.00 $\pm$ 0.00 & \textbf{100.00} $\pm$ 0.00 & 57.25 $\pm$ 0.00 & \textbf{4.59} $\pm$ 0.00 \\
\shortstack{S5$\rightarrow$T0\\(poor)} & 67.20 $\pm$ 5.22 & 98.60 $\pm$ 0.55 & 78.20 $\pm$ 4.44 & 3.98 $\pm$ 0.20 \\
\hline
\end{tabular*}
\par\vspace{2pt}
{\footnotesize\raggedright\textit{Note:} Reduction contribution sums observed per-scene $1-n_{\mathrm{out}}/n_{\mathrm{in}}$. Denominators: $^{\dagger}$100 scenes, $^{\ddagger}$510 goals.\par}
\end{table}

\begin{figure}[!t]
\centering
\includegraphics[width=\columnwidth]{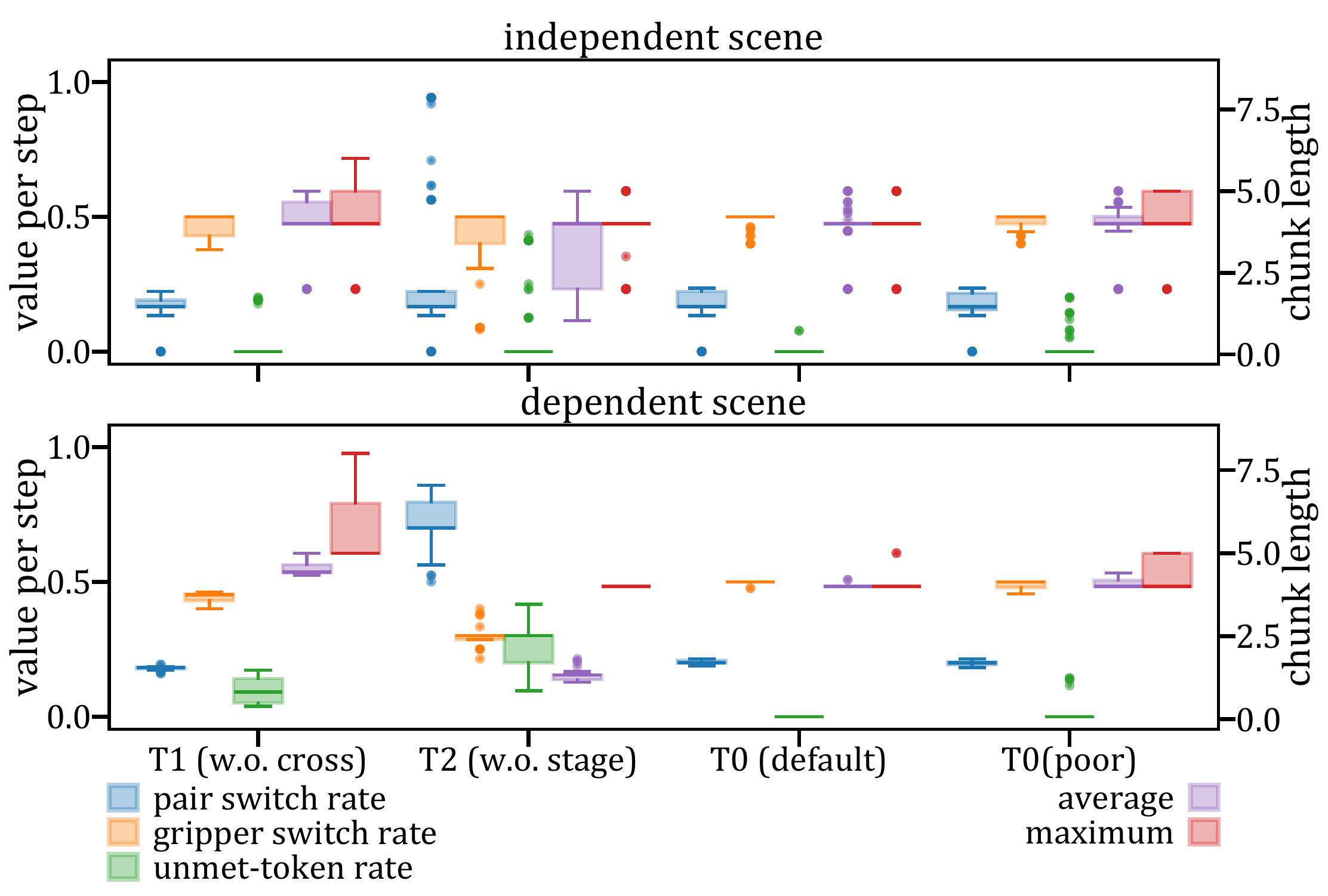}
\caption{Execution morphology of T0 and its ablations. The left axis shows switch and unmet-token rates, and the right axis shows chunk lengths. Boxes include available outputs only: T0/T1/T2/T0(poor) = 100/100/100/99. S0 uses seed 101, and each input condition uses its first generation run.}
\label{fig:morphology_boxplots}
\end{figure}

Fig.~\ref{fig:morphology_boxplots} compares sequence organization and unmet preconditions in independent and dependent scenes. A chunk comprises consecutive actions belonging to the same
assembly pair. Unmet-token rates are mostly zero in independent scenes. In dependent scenes, T0 maintains zero unmet-token rates and typically four-action chunks. T2 shows more frequent pair switching and shorter chunks, accompanied by unmet preconditions despite fewer gripper switches. This supports staged sorting in limiting action interleaving that disrupts shared gripper state. T1 exhibits longer chunks and lower switch rates but more unmet preconditions. Even with A2 inputs, T0 (poor) retains typical four-action chunks and zero median unmet-token rates, supporting TOPO's ability to maintain local action grouping.

\subsubsection{Summary of Ablation Insights}
The ablations show that a compact MLP using explicit semantic differences achieves accurate supportive-action prediction with limited supervision and low variability across training seeds. These findings support this compact design, with separate basic and supportive reasoning, cross-pair dependency inference, and staged topological sorting contributing to complete-scene success.

\subsection{Real-Robot Assembly Validation}

Assembly performance of the framework was evaluated on a physical UR3 collaborative platform with a RealSense vision testbed. The mating parts had a maximum radial clearance of 0.5~mm. Thirty challenging assembly trials incorporating occlusion and pose-deviation interference were conducted in each of two groups: scenes without and with significant cross-pair dependencies. Each group was partitioned into three equal subgroups (10 LLM-pruning-dominant trials, 10 supportive-action-dominant trials, and 10 mixed edge-case trials) to assess stability under varying task structures and interference types.

The system achieved 30/30 (100\%) success in independent scenes and 27/30 (90\%) in dependent scenes, yielding 57/60 (95\%) overall. The latter subgroup results were 10/10, 8/10, and 9/10, respectively. All three failures manifested during physical execution rather than during upstream neuro-symbolic planning. Analysis of low-level anomaly logs reveals:

\begin{itemize}
\item \textbf{Physical jamming and wedging:} Two failures during precision insertion triggered torque-threshold safety interrupts. F/T observations and simulation/physical experience suggest sim-to-real friction cone discrepancies: the PPO policy exploits idealized friction surfaces in simulation, while local surface roughness on real components narrows the friction cone, leading to lodged states.
\item \textbf{Environmental topological mutation:} In a dependent scene, removal of a stacked interfering part caused the underlying workpiece to topple an adjacent component, inducing a global topological mutation that exceeds the local correction range of atomic skills. This highlights the limitation of planning from a single RGB-D scene without real-time perceptual feedback.
\end{itemize}

\begin{figure}[!t]
\centering
\includegraphics[width=\columnwidth]{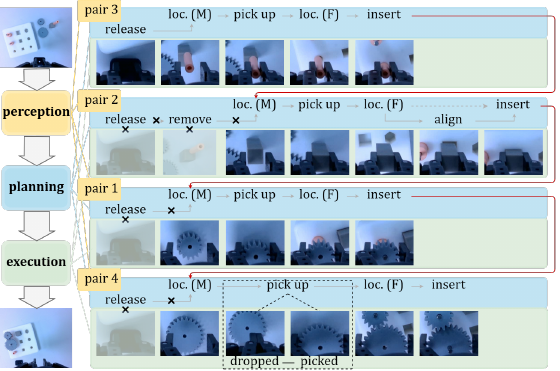}
\caption{The perception-planning-execution pipeline. Given a one-shot assembly scene, the perception layer (yellow) extracts semantic information. The planning layer (blue) generates basic local DAGs and then augments them for edge cases. Topological sorting yields a global sequence for real-world execution (green).}
\label{fig:real_assembly_group}
\end{figure}

Real-robot results align with offline trends: the system maintains high success rates in both independent and dependent scenes, with three failures observed during physical execution. These results demonstrate the integrated framework's practical deployability in the tested precision assembly tasks, as shown in Fig.~\ref{fig:real_assembly_group}.

\setlength{\textfloatsep}{1.55\baselineskip plus 0.2\baselineskip minus 0.4\baselineskip}
\setlength{\floatsep}{0.85\baselineskip plus 0.2\baselineskip minus 0.2\baselineskip}
\setlength{\dbltextfloatsep}{1.55\baselineskip plus 0.2\baselineskip minus 0.4\baselineskip}
\section{Discussion}

The neuro-symbolic framework integrates perception, planning, and execution for multi-pair assembly tasks.

\textit{Semantic Abstraction and Supportive-Action Extension:} By mapping visually different tasks to common semantic groups, the framework allows shared supportive-action requirements to be represented across objects and layouts, despite differences in geometric appearance. This abstraction supports reuse of learned knowledge across assembly structures. Separating supportive reasoning from basic-action construction also allows new actions to be introduced through additional semantics and supervision. This separation offers a basis for composing more complex supportive sub-processes from basic action primitives.

\textit{Local Reasoning and Global Coordination:} The HTN comparison shows that alternative basic-action constructors can achieve the same complete-scene success rate with shared augmentation and coordination stages. Local reasoning addresses individual task requirements, while global coordination resolves cross-pair ordering constraints and removes redundant actions. The ablations support the contribution of this coordination to complete-scene success, while the robot trials demonstrate integration with force-aware execution. However, coordination cannot supply all missing local requirements. The five failed C0 scenes illustrate this boundary: with incorrect semantic estimates from upstream, JS-Aug omitted required actions.

\textit{Future Closed-Loop Assembly:} Force-aware skills use feedback within individual actions, while changes in object relations can invalidate the remaining assembly plan. Updating scene observations after relevant execution events and replanning unfinished tasks would connect local contact adaptation with global task adaptation. Such a perception--planning--execution cycle offers a direction for longer-horizon assembly, with human--robot collaboration and shared control as further extensions.

\section{Conclusion}
We present a hierarchical neuro-symbolic framework for multi-pair robotic assembly that separates basic-action construction, supportive-action reasoning, and cross-pair coordination. Semantic differences guide supportive-action discrimination, while behavior trees connect coordinated plans to force-aware execution. In the evaluated scenes, the framework matches the tested HTN route in complete-scene success without a hand-authored HTN decomposition library and achieves higher success rates than the other tested routes. Real-robot experiments further demonstrate high assembly success in scenes with and without cross-pair dependencies.

The framework combines neural semantic reasoning with explicit symbolic constraint handling, supporting knowledge reuse and supervised extension to new actions. Its modular cache-replay protocol enables controlled evaluation of component contributions, providing a reproducible and attributable benchmark in future neuro-symbolic assembly research. Together, these capabilities provide a foundation for developing reliable and flexible autonomous industrial assembly systems. Future work will couple scene updates with replanning during execution and explore alternative force-aware execution strategies, extending planning from an initial scene toward closed-loop multi-pair assembly.



\begin{thebibliography}{10}
\providecommand{\url}[1]{#1}
\csname url@samestyle\endcsname
\providecommand{\newblock}{\relax}
\providecommand{\bibinfo}[2]{#2}
\providecommand{\BIBentrySTDinterwordspacing}{\spaceskip=0pt\relax}
\providecommand{\BIBentryALTinterwordstretchfactor}{4}
\providecommand{\BIBentryALTinterwordspacing}{\spaceskip=\fontdimen2\font plus
\BIBentryALTinterwordstretchfactor\fontdimen3\font minus
  \fontdimen4\font\relax}
\providecommand{\BIBforeignlanguage}[2]{{%
\expandafter\ifx\csname l@#1\endcsname\relax
\typeout{** WARNING: IEEEtran.bst: No hyphenation pattern has been}%
\typeout{** loaded for the language `#1'. Using the pattern for}%
\typeout{** the default language instead.}%
\else
\language=\csname l@#1\endcsname
\fi
#2}}
\providecommand{\BIBdecl}{\relax}
\BIBdecl

\bibitem{ranjan_das_toward_2025}
A.~Ranjan~Das and M.~Koskinopoulou, ``Toward sustainable manufacturing: A
  review on innovations in robotic assembly and disassembly,'' \emph{IEEE
  Access}, vol.~13, pp. 100\,149--100\,166, 2025.

\bibitem{karbouj_adaptive_2026}
B.~Karbouj, R.~Garha, K.~Ke{\ss}ler, and J.~Kr{\"u}ger, ``Adaptive robotic
  behavior in industrial human--robot collaboration: A systematic review of
  taxonomies, enabling mechanisms, and research frontiers,'' \emph{IEEE
  Access}, vol.~14, pp. 1398--1422, 2026.

\bibitem{wang_llm-based_2025}
B.~Wang, L.~Zheng, Y.~Wang, and Z.~Qi, ``{{LLM-based}} multi-agent task
  planning for human-robot collaborative assembly balancing operator experience
  and efficiency,'' \emph{J. Manuf. Syst.}, vol.~82, pp. 1020--1045, Oct. 2025.

\bibitem{zhang_surrogate_2025}
Y.~Zhang, W.~Luo, Y.~Hu, Q.~Wang, L.~Cheng, and Y.~Ke, ``A surrogate
  model-driven assembly coordination framework for aircraft components based on
  cooperative multi-agent deep reinforcement learning,'' \emph{J. Manuf.
  Syst.}, vol.~83, pp. 46--64, Dec. 2025.

\bibitem{van_wyk_comparative_2018}
K.~Van~Wyk, M.~Culleton, J.~Falco, and K.~Kelly, ``Comparative peg-in-hole
  testing of a force-based manipulation controlled robotic hand,'' \emph{IEEE
  Trans. Robot.}, vol.~34, no.~2, pp. 542--549, Apr. 2018.

\bibitem{fox_pddl21_2003}
M.~Fox and D.~Long, ``{PDDL2}.1: An extension to {PDDL} for expressing temporal
  planning domains,'' \emph{J. Artif. Intell. Res.}, vol.~20, pp. 61--124, Dec.
  2003.

\bibitem{nau_shop2_2003}
D.~S. Nau \emph{et~al.}, ``{SHOP2}: An {HTN} planning system,'' \emph{J. Artif.
  Intell. Res.}, vol.~20, pp. 379--404, Dec. 2003.

\bibitem{pan_task_2024}
T.~Pan, R.~Shome, and L.~E. Kavraki, ``Task and motion planning for execution
  in the real,'' \emph{IEEE Trans. Robot.}, vol.~40, pp. 3356--3371, 2024.

\bibitem{liu_delta_2025}
Y.~Liu, L.~Palmieri, S.~Koch, I.~Georgievski, and M.~Aiello, ``{DELTA}:
  Decomposed efficient long-term robot task planning using large language
  models,'' in \emph{Proc. IEEE Int. Conf. Robot. Autom. (ICRA)}, May 2025, pp.
  10\,995--11\,001.

\bibitem{sun_neurosymbolic_2024}
X.~Sun and Y.~Shoukry, ``Neurosymbolic motion and task planning for linear
  temporal logic tasks,'' \emph{IEEE Trans. Robot.}, vol.~40, pp. 2749--2768,
  2024.

\bibitem{ao_llm-as-bt-planner_2025}
J.~Ao, F.~Wu, Y.~Wu, A.~Swiki, and S.~Haddadin, ``{LLM-as-BT-Planner}:
  Leveraging {LLMs} for behavior tree generation in robot task planning,'' in
  \emph{Proc. IEEE Int. Conf. Robot. Autom. (ICRA)}, May 2025, pp. 1233--1239.

\bibitem{li_grhp_2026}
X.~Li, G.~Tian, Y.~Cui, X.~Shao, and Z.~Wang, ``{GRHP}: Graph-fused
  hierarchical planning for embodied long-horizon robotic task,'' \emph{Eng.
  Appl. Artif. Intell.}, vol. 165, Feb. 2026, {Art. no.} 113413.

\bibitem{homem_de_mello_andor_1990}
L.~S. {Homem de Mello} and A.~C. Sanderson, ``{{AND/OR}} graph representation
  of assembly plans,'' \emph{IEEE Trans. Robot. Autom.}, vol.~6, no.~2, pp.
  188--199, Apr. 1990.

\bibitem{jimenez_survey_2013}
P.~Jim{\'e}nez, ``Survey on assembly sequencing: A combinatorial and
  geometrical perspective,'' \emph{J. Intell. Manuf.}, vol.~24, no.~2, pp.
  235--250, Apr. 2013.

\bibitem{keramat_decentralized_2026}
F.~Keramat, S.~Salimi, and T.~Westerlund, ``Decentralized intent-based
  multi-robot task planner with {LLM} oracles on {Hyperledger Fabric},'' 2026,
  arXiv:2602.08421.

\bibitem{gottardi_had-tamp_2026}
A.~Gottardi, M.~Terreran, E.~Pagello, and E.~Menegatti, ``{HAD-TAMP}: Human
  adaptive task and motion planning for human--robot collaboration in
  industrial scenario,'' \emph{Robot. Auton. Syst.}, vol. 198, Apr. 2026, {Art.
  no.} 105318.

\bibitem{chen_behavioral_2026}
B.~Chen \emph{et~al.}, ``Behavioral planning and parameter meta learning for
  embodied intelligence robots in adaptive assembly,'' \emph{J. Ind. Inf.
  Integr.}, vol.~49, Jan. 2026, {Art. no.} 100995.

\bibitem{kawaharazuka_vision-language-action_2025}
K.~Kawaharazuka, J.~Oh, J.~Yamada, I.~Posner, and Y.~Zhu,
  ``Vision-language-action models for robotics: A review towards real-world
  applications,'' \emph{IEEE Access}, vol.~13, pp. 162\,467--162\,504, 2025.

\bibitem{liu_vla-pruner_2026}
Z.~Liu \emph{et~al.}, ``{VLA-Pruner}: Temporal-aware dual-level visual token
  pruning for efficient vision-language-action inference,'' 2026,
  arXiv:2511.16449v3.

\bibitem{wang_vlabot_2026}
X.~Wang, X.~Dengxiong, S.~Bai, P.~Zheng, and Y.~Zhang, ``{{VLAbot}}: A human
  vision--language--action models interaction framework for robotic assembly,''
  \emph{Robot. Comput.-Integr. Manuf.}, vol. 100, Aug. 2026, {Art. no.} 103268.

\bibitem{tang_vlm-dewm_2026}
G.~Tang \emph{et~al.}, ``{VLM-DEWM}: Dynamic external world model for
  verifiable and resilient vision-language planning in manufacturing,'' 2026,
  arXiv:2602.15549.

\bibitem{chen_generalizable_2026}
H.~Chen, T.~Kiyokawa, W.~Wan, and K.~Harada, ``Generalizable task-oriented
  object grasping through {{LLM-guided}} ontology and similarity-based
  planning,'' \emph{Robot. Auton. Syst.}, vol. 201, Jul. 2026, {Art. no.}
  105442.

\bibitem{zhou_genco_2025}
Z.~Zhou, S.~Veeramani, H.~Fakhruldeen, S.~Uyanik, and A.~I. Cooper,
  ``{{GenCo}}: A dual {{VLM}} generate-correct framework for adaptive
  peg-in-hole robotics,'' in \emph{Proc. IEEE Int. Conf. Robot. Autom. (ICRA)},
  May 2025, pp. 16\,744--16\,751.

\bibitem{zhou_human---loop_2025}
H.~Zhou, Y.~Lin, L.~Yan, and H.~Min, ``Human-in-the-loop learning for adaptive
  robot manipulation using large language models and behavior trees,'' in
  \emph{Proc. IEEE/RSJ Int. Conf. Intell. Robots Syst. (IROS)}, Oct. 2025, pp.
  19\,039--19\,046.

\bibitem{garrett_integrated_2021}
C.~R. Garrett \emph{et~al.}, ``Integrated task and motion planning,''
  \emph{Annu. Rev. Control Robot. Auton. Syst.}, vol.~4, no.~1, pp. 265--293,
  May 2021.

\bibitem{mcallester_systematic_1991}
\BIBentryALTinterwordspacing
D.~McAllester and D.~Rosenblitt, ``Systematic nonlinear planning,'' in
  \emph{Proc. 9th Nat. Conf. Artif. Intell. (AAAI)}, 1991, pp. 634--639.
  [Online]. Available: \url{https://cdn.aaai.org/AAAI/1991/AAAI91-099.pdf}
\BIBentrySTDinterwordspacing

\bibitem{penberthy_ucpop_1992}
\BIBentryALTinterwordspacing
J.~S. Penberthy and D.~S. Weld, ``{{UCPOP}}: A sound, complete, partial order
  planner for {{ADL}},'' in \emph{Proc. 3rd Int. Conf. Princ. Knowl. Represent.
  Reason. (KR)}, 1992, pp. 103--114. [Online]. Available:
  \url{https://homes.cs.washington.edu/~weld/papers/ucpop-kr92.pdf}
\BIBentrySTDinterwordspacing

\bibitem{zhou_llm-bt_2024}
H.~Zhou, Y.~Lin, L.~Yan, J.~Zhu, and H.~Min, ``{LLM-BT}: Performing robotic
  adaptive tasks based on large language models and behavior trees,'' in
  \emph{Proc. IEEE Int. Conf. Robot. Autom. (ICRA)}, May 2024, pp.
  16\,655--16\,661.

\bibitem{styrud_automatic_2025}
J.~Styrud, M.~Iovino, M.~Norrl{\"o}f, M.~Bj{\"o}rkman, and C.~Smith,
  ``Automatic behavior tree expansion with {{LLMs}} for robotic manipulation,''
  in \emph{Proc. IEEE Int. Conf. Robot. Autom. (ICRA)}, May 2025, pp.
  1225--1232.

\bibitem{izzo_btgenbot_2024}
R.~A. Izzo, G.~Bardaro, and M.~Matteucci, ``{BTGenBot}: Behavior tree
  generation for robotic tasks with lightweight {LLMs},'' in \emph{Proc.
  IEEE/RSJ Int. Conf. Intell. Robots Syst. (IROS)}, Oct. 2024, pp. 9684--9690.

\bibitem{cai_hbtp_2025}
Y.~Cai \emph{et~al.}, ``{HBTP}: Heuristic behavior tree planning with large
  language model reasoning,'' in \emph{Proc. IEEE Int. Conf. Robot. Autom.
  (ICRA)}, May 2025, pp. 13\,706--13\,713.

\bibitem{kwon_fast_2025}
M.~Kwon, Y.~Kim, and Y.~J. Kim, ``Fast and accurate task planning using
  neuro-symbolic language models and multi-level goal decomposition,'' in
  \emph{Proc. IEEE Int. Conf. Robot. Autom. (ICRA)}, May 2025, pp.
  16\,195--16\,201.

\bibitem{merlo_exploiting_2025}
E.~Merlo, M.~Lagomarsino, E.~Lamon, and A.~Ajoudani, ``Exploiting information
  theory for intuitive robot programming of manual activities,'' \emph{IEEE
  Trans. Robot.}, vol.~41, pp. 1245--1262, 2025.

\bibitem{ichter_do_2023}
\BIBentryALTinterwordspacing
B.~Ichter \emph{et~al.}, ``Do as {I} can, not as {I} say: Grounding language in
  robotic affordances,'' in \emph{Proc. 6th Conf. Robot Learn. (CoRL)}, 2023,
  pp. 287--318. [Online]. Available:
  \url{https://proceedings.mlr.press/v205/ichter23a.html}
\BIBentrySTDinterwordspacing

\bibitem{singh_progprompt_2023}
I.~Singh \emph{et~al.}, ``{{ProgPrompt}}: Generating situated robot task plans
  using large language models,'' in \emph{Proc. IEEE Int. Conf. Robot. Autom.
  (ICRA)}, 2023, pp. 11\,523--11\,530.

\bibitem{liang_code_2023}
J.~Liang \emph{et~al.}, ``Code as policies: Language model programs for
  embodied control,'' in \emph{Proc. IEEE Int. Conf. Robot. Autom. (ICRA)},
  2023, pp. 9493--9500.

\bibitem{liu_llmp_2023}
B.~Liu \emph{et~al.}, ``{{LLM+P}}: Empowering large language models with
  optimal planning proficiency,'' 2023, arXiv:2304.11477.

\bibitem{kambhampati_llm_modulo_2024}
\BIBentryALTinterwordspacing
S.~Kambhampati \emph{et~al.}, ``Position: {{LLMs}} can't plan, but can help
  planning in {{LLM-Modulo}} frameworks,'' in \emph{Proc. Int. Conf. Mach.
  Learn. (ICML)}, 2024, pp. 22\,895--22\,907. [Online]. Available:
  \url{https://proceedings.mlr.press/v235/kambhampati24a.html}
\BIBentrySTDinterwordspacing

\bibitem{mendez-mendez_systematic_2025}
J.~{Mendez-Mendez}, ``A systematic study of large language models for task and
  motion planning with {{PDDLStream}},'' 2025, arXiv:2510.00182v1.

\bibitem{shao_breaking_2025}
J.-J. Shao, H.-J. You, G.~Cai, Q.~Dai, Z.~Dong, and L.-Z. Guo, ``Breaking the
  self-evaluation barrier: Reinforced neuro-symbolic planning with large
  language models,'' in \emph{Proc. 34th Int. Joint Conf. Artif. Intell.
  (IJCAI)}, Aug. 2025, pp. 6129--6137.

\bibitem{bhat_grounding_nodate}
V.~Bhat, A.~U. Kaypak, P.~Krishnamurthy, R.~Karri, and F.~Khorrami, ``Grounding
  large language models for robot task planning using closed-loop state
  feedback,'' \emph{Adv. Robot. Res.}, Nov.~10, 2025, early access.

\bibitem{rodriguez-guerra_deliberative_2025}
D.~{Rodriguez-Guerra}, O.~Avram, S.~Baraldo, M.~Zamboni, and A.~Valente,
  ``Deliberative layered behavior tree approach for real-time concurrent
  decision-making in human-robot interaction for assembly,'' in \emph{Proc.
  IEEE Int. Conf. Robot Hum. Interact. Commun. (RO-MAN)}, Aug. 2025, pp.
  231--236.

\bibitem{galitsky_neuro-symbolic_2026}
B.~Galitsky and A.~Rybalov, ``Neuro-symbolic verification for preventing
  {{LLM}} hallucinations in process control,'' \emph{Processes}, vol.~14,
  no.~2, Jan. 2026, {Art. no.} 322.

\bibitem{parnada_towards_2026}
A.~Parnada, M.~Qu, M.~Castellani, H.~J. Chang, and Y.~Wang, ``Towards
  cost-effective and safe contact-rich robotic manipulation with reinforcement
  learning: A review of techniques for future industrial automation,''
  \emph{Proc. Inst. Mech. Eng. I, J. Syst. Control Eng.}, vol. 240, no.~1, pp.
  3--35, Jan. 2026.

\bibitem{xie_towards_2025}
W.~Xie and N.~Correll, ``Towards forceful robotic foundation models: A
  literature survey,'' 2025, arXiv:2504.11827.

\bibitem{liao_friction-aware_2026}
Z.~Liao, C.~Gong, F.~Zhao, and X.~Mei, ``Friction-aware robotic peg-in-hole
  assembly via contact wrench measurement,'' \emph{Measurement}, vol. 257, Jan.
  2026, {Art. no.} 118781.

\bibitem{tracy_efficient_2025}
K.~Tracy \emph{et~al.}, ``Efficient online learning of contact force models for
  connector insertion,'' in \emph{Proc. IEEE Int. Conf. Robot. Autom. (ICRA)},
  May 2025, pp. 10\,944--10\,950.

\bibitem{hou_force-based_2025}
J.~Hou, S.~Xing, Y.~Ma, and F.~Jing, ``Force-based cabin insertion assembly
  with an off-robot force sensing configuration,'' \emph{IFAC-PapersOnLine},
  vol.~59, no.~35, pp. 472--477, 2025.

\bibitem{shen_learning-based_2025}
Z.~Shen, Z.~Jiang, J.~Zhang, J.~Wu, and Q.~Zhu, ``Learning-based robot assembly
  method for peg insertion tasks on inclined hole using time-series force
  information,'' \emph{Biomim. Intell. Robot.}, vol.~5, no.~1, Mar. 2025, {Art.
  no.} 100209.

\bibitem{ding_ensemble_2026}
G.~Ding, X.~Zang, P.~Wang, X.~Zhang, Y.~Zhu, and J.~Zhao, ``An ensemble
  reinforcement learning framework for robotic high-precision peg-in-hole
  assembly via human demonstrations,'' \emph{Robot. Comput.-Integr. Manuf.},
  vol. 100, Aug. 2026, {Art. no.} 103279.

\bibitem{shirai_sim--real_2025}
\BIBentryALTinterwordspacing
Y.~Shirai, K.~Ota, D.~K. Jha, and D.~Romeres, ``Sim-to-real contact-rich
  pivoting via optimization-guided {{RL}} with vision and touch,'' in
  \emph{Proc. NeurIPS Workshop Embodied World Models for Decision Making}, Dec.
  2025. [Online]. Available: \url{https://www.merl.com/publications/TR2025-169}
\BIBentrySTDinterwordspacing

\bibitem{stranghoner_share-rl_2026}
J.~Strangh{\"o}ner, P.~Hartmann, L.-M. Weigelt, M.~Braun, S.~Wrede, and
  K.~Neumann, ``{SHaRe-RL}: Structured, interactive reinforcement learning for
  contact-rich industrial assembly tasks,'' in \emph{Proc. IEEE Int. Conf.
  Robot. Autom. (ICRA)}, Jun. 2026, pp. 1759--1766.

\bibitem{he_mask_2017}
K.~He, G.~Gkioxari, P.~Doll{\'a}r, and R.~Girshick, ``Mask {{R-CNN}},'' in
  \emph{Proc. IEEE Int. Conf. Comput. Vis. (ICCV)}, Oct. 2017, pp. 2980--2988.

\bibitem{radford_learning_2021}
\BIBentryALTinterwordspacing
A.~Radford \emph{et~al.}, ``Learning transferable visual models from natural
  language supervision,'' in \emph{Proc. Int. Conf. Mach. Learn. (ICML)}, Jul.
  2021, pp. 8748--8763. [Online]. Available:
  \url{https://proceedings.mlr.press/v139/radford21a.html}
\BIBentrySTDinterwordspacing

\bibitem{todorov_mujoco_2012}
E.~Todorov, T.~Erez, and Y.~Tassa, ``{MuJoCo}: A physics engine for model-based
  control,'' in \emph{Proc. IEEE/RSJ Int. Conf. Intell. Robots Syst. (IROS)},
  Oct. 2012, pp. 5026--5033.

\end{thebibliography}
\end{document}